\documentclass{article}

\usepackage{microtype}
\usepackage{graphicx}
\usepackage{subcaption}
\usepackage{booktabs} % for professional tables

\usepackage{hyperref}

\usepackage[preprint]{icml2026}

\usepackage{amsmath}
\usepackage{amssymb}
\usepackage{mathtools}
\usepackage{amsthm}
\usepackage{pifont} % For check and cross symbols
\usepackage[utf8]{inputenc} % allow utf-8 input
\usepackage[T1]{fontenc}    % use 8-bit T1 fonts
\usepackage{xcolor}
\usepackage{booktabs}
\usepackage{wrapfig}   % in the preamble
\usepackage{enumitem}
\usepackage{booktabs}

\definecolor{pastelblue}{RGB}{90, 140, 200}   % Slightly darker blue
\definecolor{pastelgreen}{RGB}{90, 180, 90}   % Slightly darker green
\definecolor{pastelred}{RGB}{200, 100, 100}   % Sli
\definecolor{darkgreen}{HTML}{2CA02C}
\newcommand{\cmark}{\textcolor{green!60!black}{\ding{51}}} % check mark
\newcommand{\xmark}{\textcolor{red}{\ding{55}}} % cross mark
\newcommand{\E}{\mathbb{E}}
\usepackage[capitalize,noabbrev]{cleveref}

\theoremstyle{plain}

\theoremstyle{definition}

\theoremstyle{remark}

\usepackage[textsize=tiny]{todonotes}

\icmltitlerunning{Performance Asymmetry in Model-Based Reinforcement Learning}

\begin{document}

\twocolumn[
  \icmltitle{Performance Asymmetry in Model-Based Reinforcement Learning}



  \icmlsetsymbol{equal}{*}

  \begin{icmlauthorlist}
    \icmlauthor{Jing Yu Lim}{nus}
    \icmlauthor{Rushi Sha}{nus}
    \icmlauthor{Zarif Ikram}{nus}
    \icmlauthor{Samson Yu Bai Jian}{nus}
    \icmlauthor{Haozhe Ma}{nus}
    \icmlauthor{Tze-Yun Leong}{nus}
    \icmlauthor{Dianbo Liu}{nus}
  \end{icmlauthorlist}

  \icmlaffiliation{nus}{National University of Singapore}
  % \icmlaffiliation{med}{Yong Loo Lin School of Medicine, National University of Singapore}

  \icmlcorrespondingauthor{Jing Yu Lim}{jing.yu@nus.edu.sg}
  \icmlcorrespondingauthor{Dianbo Liu}{dianbo@nus.edu.sg}
  \icmlcorrespondingauthor{Tze-Yun Leong}{dianbo@nus.edu.sg}

  \icmlkeywords{Machine Learning, ICML}

  \vskip 0.3in
]



\printAffiliationsAndNotice{}  % no special notice (required even if empty)

% driven by reinforcement learning agents trained on powerful diffusion world models

% This is especially pronounced in pixel-based agents trained with diffusion world models.

% We address the problematic aggregates by delineating all tasks as Agent-Optimal or Human-Optimal and advocate for equal importance on metrics from both sets

% this pronounced asymmetry is due to the lack of temporally-structured latent space trained with the World Model objective in pixel-based methods. Lastly, to address this issue, 

% , and runs 3 times faster with 43\% lower memory than the latest pixel-based diffusion baseline. Overall, our work rethinks what it truly means to cross human-level performance in Atari100k.
% advocate for the use the harmonic mean of the overall performance across the two sets of tasks (Sym-HNS)
% the exceptionally high scores in a handful of Agent-Optimal tasks mask and overcompensate for the failures in many Human-Optimal tasks. 
% . the exceptional outperformance in Agent-Optimal tasks conceals the failures in Human-Optimal tasks
% Using the standard aggregate of mean Human-Normalized Scores (HNS) results in the former overcompensating for the failures in the latter; indeed,  
\begin{abstract}
  Recently, Model-Based Reinforcement Learning (MBRL) have achieved average super-human level performance on the Atari100k benchmark. However, we discover that conventional aggregates mask a major problem, $\textbf{Performance Asymmetry}$: MBRL agents dramatically outperform humans in certain tasks (Agent-Optimal tasks) while drastically underperform humans in other tasks (Human-Optimal tasks). Indeed, \textbf{despite achieving SOTA in the overall mean Human-Normalized Scores (HNS), the SOTA agent scored the worst among baselines on Human-Optimal tasks, with a striking 21$\times$ performance gap between the Human/Agent-Optimal subsets}. To address this, we partition Atari100k evenly into Human/Agent-Optimal subsets, and introduce a more balanced aggregate, $\textbf{Sym-HNS}$. We further trace the striking Performance Asymmetry in the SOTA pixel diffusion world model to the curse of dimensionality and its prowess on high visual detail tasks (e.g. $\textit{Breakout}$). To this end, we propose a novel latent end-to-end \textbf{Joint Embedding DIffusion (JEDI)} world model that achieves SOTA results in Sym-HNS, Human-Optimal tasks, and $\textit{Breakout}$—thus \textbf{reversing the worsening Performance Asymmetry trend} while improving computational efficiency and remaining competitive on the full Atari100k.
\end{abstract}
  % To verify this, by showing that the latent-based agents consistently outperform pixel-based agents on Sym-HNS
\section{Introduction}
\label{sec:intro}
% Model-based reinforcement learning (MBRL) uses \textit{world models} \citep{sutton1991DYNA,ha2018World} to learn the environment dynamics to enhance decision-making efficiency and performance in sequential decision tasks. Unlike model-free approaches that directly approximate value functions or policies from interactions, MBRL learns an internal representation of the environment to anticipate outcomes and plan ahead. This enables the agent to plan and reason ahead, improving sample efficiency and adaptability in complex or partially observable environments \citep{Hafner2020Dreamer, hafner2020Dreamerv2, wu2023daydreamer, hafner2023Dreamerv3}. 

 Model-Based Reinforcement Learning (MBRL) centers on learning a model of the agent's  environment dynamics known as a \textit{World Model} \cite{sutton1991DYNA,ha2018World}. The principle accompanying MBRL is rooted in broader theories of intelligence; \cite{YiMa2022PrinciplesIntelligence} asserts \textit{parsimony} and \textit{self-consistency} as foundational for the emergence of intelligence---a view shared by \citet{lecun2022path} and echoes earlier works by \citet{friston2006FEP}. In the MBRL paradigm introduced by \citet{Hafner2020Dreamer}, parsimony is achieved by compressing high-dimensional images to a low-dimensional latent space using the encoder, and it is partly learned through the self-consistency objective of aligning the trajectories generated by its dynamics model with actual observations.
 % It is also aligned with the \textit{Free Energy Principle} \citep{friston2006FEP, friston2010FEP}, which argues that intelligence emerges from surprise minimization between an agent's internal predictions and its observations.
 % This setup forms an iterative, coupled feedback loop with a Reinforcement Learning (RL) agent: the world model improves by maximizing self-consistency between its dynamics model predictions and its observations, while the RL policy learns using the imagined latent trajectories. The improved then collects higher-quality, further enhancing world-model learning and the cycle repeats.

This paradigm is effective, recently crossing the mean human-level performance \citep{ micheli2022iris, robine2023TWM, hafner2023Dreamerv3,zhang2023storm} on the challenging Atari100k benchmark \citep{kaiser2019Atari100kSimPLE}. Since Atari100k limits the agent to 100K environment steps or around two hours of human gameplay, this is a seminal milestone for RL as one of its core limitations was poor data efficiency. However, we discover that \textit{crossing mean human-level performance} with MBRL on Atari100k can be problematic.
% , a stark contrast to pre-existing benchmarks with hundreds of million environment steps \citep{bellemare2013ALE}, 
% TheDespite crossing Unfortunately,evel performance is a notable milestone. When the seminal model-free RL work by \cite{mnih2015HumanLevelAtari} crossed the barrier in 2015, it scored close to a professional game tester on the majority of the tasks, sparking a deep RL revolution that led to breakthroughs like \cite{silver2016AlphaGo, silver2017MuZero}.
% \begin{figure}[t]
%     \centering
%     \includegraphics[width=\textwidth]{pdfs/T_plot_all_results.pdf}
%     \caption{Performance asymmetry. There exist a large disparity in performance of MBRL agents between Agent-Optimal and Human-optimal games. MBRL agents are better in Agent-Optimal games by at least an order of magnitude. This skew in performance is even more pronounced in pixel-based MBRL agents.}
%     \label{fig:T_plot_all_results}    
% \end{figure}

\begin{figure*}[t]
  \centering
  % \hfill
  % % First subfigure
  % \begin{subfigure}[c]{0.18\textwidth}
  %   \centering
  %   \includegraphics[width=\textwidth]{pdfs/fig1_starwars_cartoon.pdf}
  % \end{subfigure}
  % \hfill
  % \vrule
  % \hfill
  % \begin{subfigure}[c]{0.78\textwidth}
    % \centering
  \includegraphics[width=\textwidth]{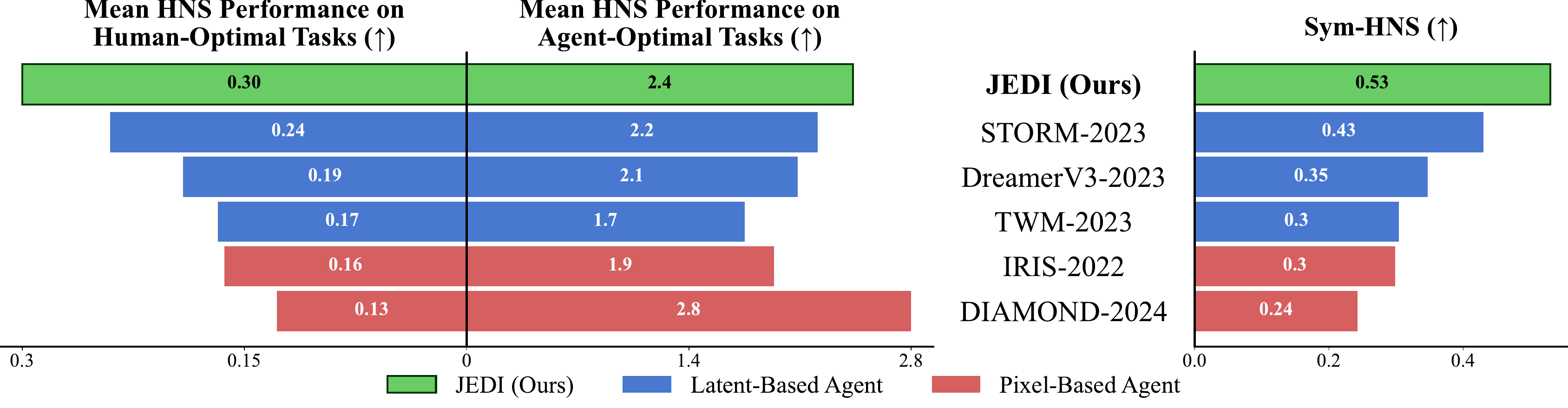}
  % \end{subfigure}
  \caption{\textbf{Left:} MBRL agents exhibit a large Performance Asymmetry between Human-Optimal and Agent-optimal tasks (defined in \cref{sec:performance_bias}), often outperforming in the latter by over an order of magnitude. This disparity is especially striking in pixel-based MBRL agents, with \citet{alonso2024diamond} exhibiting a 21 $\times$ disparity and performing worst on Human-Optimal tasks despite being SOTA overall. \textbf{Right:} Sym-HNS is our proposed metric which uses the harmonic mean for balanced aggregation. With end-to-end latent diffusion dynamics model, JEDI reverses the worsening Performance Asymmetry trend and achieves SOTA Sym-HNS.} 
  \label{fig:T_plot_all_results}
\end{figure*}
 % Performance Asymmetry illustration: tasks where humans excel often differ from RL agents \footnotemark. 
 % , similar to how F1 score balances two equally important metrics
% \footnotetext{This can be due to reward hacking or issues with representation learning~\citep{skalse2022RewardHack, guo2021machine}}
In particular, we uncover a surprising \textbf{Performance Asymmetry} in MBRL performance across the 26 tasks in Atari100k. While the mean Human Normalized Score (HNS)\footnote{HNS = $\frac{\textnormal{Agent Score - Random Score}}{\textnormal{Human Score - Random Score}}$\citep{mnih2015HumanLevelAtari}} of more than $1.0$ suggests super-human performance, a closer look reveals that MBRL agents massively outperform in tasks with a low human score---what we call \textbf{\textit{Agent-Optimal tasks}}---and perform overwhelmingly poor in tasks with a high human score---\textbf{\textit{Human-Optimal tasks}}\footnote{We define the sets of agent-optimal tasks and human-optimal tasks consisting of 13 tasks each in \cref{sec:performance_bias}.}

 % For example, the recent top-performing approach uses a diffusion model \citep{song2019NCSN, ho2020ddpm} as the dynamics model enabling pixel-perfect rollout that a pixel-based RL agent trains on, driving its state-of-the-art (SOTA) performance~\citep{alonso2024diamond}. Yet, these models may overfit \textit{some} benchmark metric rather than \textit{match} human-performance.
This is clearly illustrated in Figure \ref{fig:T_plot_all_results}: MBRL algorithms generally underperform on Human-Optimal tasks by an order of magnitude compared to Agent-Optimal tasks. Worse still, a worsening trend can be observed with pixel-based agents, with the latest pixel-based method \citep{alonso2024diamond} having the most Performance Asymmetry. Indeed, \textbf{despite achieving an overall state-of-the-art (SOTA) mean HNS of $1.46$, the SOTA agent performed the worst among all baselines on Human-Optimal tasks, with a striking 21$\times$ performance gap between Human/Agent-Optimal subsets}. This raises a critical question: \textit{are MBRL agents truly progressing or are they overfitting to certain tasks?} 
%{WHY SHOULD PEOPLE CARE ABOUT PERFORMANCE ASYMMETRY? WARRANT}

% overall super-human performance can be achieved with a handful of exceptionally high scores on Agent-Optimal tasks, overcompensating for low scores on many Human-Optimal tasks
To our knowledge, this work is the first to illuminate Performance Asymmetry in MBRL agents, highlighting the issue of task overfitting. More importantly, this brings attention to the \textbf{overlooked fact that MBRL is critically failing on certain tasks while performing exceedingly well in others} (\cref{fig:mbrl_vs_human_rawscore}). The analysis of these success and failure modes empowers future research that would have otherwise been overlooked and unrecognized. 

We first lay the groundwork by identifying the \textbf{core issue of current aggregates: (i) the arithmetic mean and (ii) the large value range of HNS}. The arithmetic mean is invariant to the extreme values, and the large value range renders meaningful changes in low values overshadowed by small delta in large values. This results in the invisibility of Performance Asymmetry in many conventional aggregates, not only in the average score (\cref{sec:existing_evaluation}). To address this, we first expose Performance Asymmetry by evenly partitioning Atari100k into Agent/Human-Optimal subsets (\cref{sec:performance_bias}). We then introduce \textbf{Sym-HNS}, a more balanced aggregate that overcomes these limitations by leveraging the harmonic mean (\cref{sec:sym-hns_evaluation}).
%similar to the F1 score \citep{F1vanRijsbergen1979} balances two equally important metrics. This overcomes the limitation of arithmetic mean which conceals failures. 
% Moreover, it appears that the MBRL community has moved on from the Atari100k benchmark, with recent works focusing on other problems like open-ended worlds and offline RL \citep{bruce2024genie, parkerholder2024genie2, Hafner2025DreamerV4}
% This can be achieved through the analysis of MBRL agents' failure in  Human-Optimal tasks and their success in Agent-Optimal tasks, 
% Hence, with the discovery of Performance Asymmetry, we can unlock advancements in the field that previously would have been overlooked and unrecognized.  
% We address this problem by advocating for placing equal importance on performance metrics across both sets of tasks, hence crediting methods which advances the performance of Human-Optimal tasks as well (Section \ref{sec:fairness_evaluation}).
 % Using the Agent/Human-Optimal subsets and Sym-HNS, we observe that pixel-based agents exhibit greater Performance Asymmetry.
 
As the initial foray into tackling this problem, we focus on reversing the worsening trend of Performance Asymmetry over time, specifically the striking asymmetry exhibited by the latest SOTA agent (Figure \ref{fig:T_plot_all_results}). We trace the underperformance of pixel-based agents in Human-Optimal tasks to the curse of dimensionality from using image inputs to its RL networks \citep{watter2015embed_Latent, oh2017value_Latent, gregor2019shaping_Latent}, coupled with the fact that Human-Optimal tasks are generally more complex (\cref{fig:action_space_shooter_comparison}). In parallel, the SOTA agents' outperformance in Agent-Optimal tasks can be attributed to their outsized performance in tasks with high visual detail (e.g. \textit{Breakout}), due to the use of a pixel diffusion world model \citep{alonso2024diamond}. As such, we seek to answer the research questions: 
\begin{enumerate}[noitemsep, topsep=0pt]
    \item Can the striking Performance Asymmetry of a pixel-based agent be mitigated with a latent space world model?
    \item Can the use of diffusion in a latent dynamics model overcome the underperformance of latent-based agents on high visual detail tasks?
\end{enumerate}
% ,where its RL networks have high-dimensional inputs instead of compressed latent states  
% Since  given the same training budget as latent-based agents, it is more difficult for pixel-based agents to learn the optimal policy, resulting in poorer performance in Human-Optimal tasks.
%By combining the visual modeling power of diffusion models with the dimensionality reduction of latent world models, JEDI bridges the gap across Agent-Optimal and Human-Optimal tasks.  
% the pixel-based policy has full information of the state from the image input. This is in contrast to latent-based agents where important visual information (e.g. position of projectile) can be lost during the compression of pixels to latents
% with the implementation of a latent space learned through diffusion world modelling
% while enjoying 3$\times$ faster inference, 2$\times$ training, and requiring only 57\% GPU memory compared to the SOTA diffusion-based world model~\citep{alonso2024diamond}. 
% and remains competitive on Agent-Optimal tasks and across the full Atari100k benchmark.
To this end, we propose the \textbf{J}oint \textbf{E}mbedding \textbf{DI}ffusion \textbf{(JEDI)} world model, a latent end-to-end diffusion model with its encoder trained using a self-consistent denoising objective inspired by Joint Embedding Predictive Architecture (JEPA) \citep{assran2023I-JEPA}. To our knowledge, \textbf{JEDI is the first method that enables the encoder to learn a latent representation that works well for the downstream RL task directly from the gradients of a latent diffusion model's denoising training loss using a JEPA-style framework}---without using reconstruction, hidden state extraction, and distillation. This is also the first work that shows that an off-the-shelf diffusion denoising loss \citep{karras2022EDM} is directly compatible with JEPA, challenging prevailing beliefs that JEPA requires bespoke loss functions~\citep{grill2020BYOL, caron2021DINO, bardes2024V-JEPA}. JEDI yields SOTA performance on Sym-HNS, Human-Optimal tasks and \textit{Breakout}---thus \textbf{reversing the worsening Performance Asymmetry trend} in MBRL while improving computational efficiency (\cref{fig:compute_and_runtime_comparison}) and remaining competitive on Agent-Optimal tasks and the full Atari100k.

In summary, our contributions are as follows (\cref{tab:claim_evidence_jedi}).
\begin{enumerate}[noitemsep, topsep=0pt]
    \item We uncover Performance Asymmetry in MBRL agents, highlighting the problem of task overfitting (\cref{sec:performance_bias}).
    \item We lay the groundwork for tackling Performance Asymmetry by partitioning Atari100k into two subsets and introducing a more balanced aggregate, Sym-HNS (\cref{sec:sym-hns_evaluation}).
    \item We analyze the striking Performance Asymmetry in pixel-based agents and propose relevant research questions (\cref{sec:agent_vs_human_task_analysis}).
    \item  We introduce JEDI, a novel latent end-to-end diffusion-based world model to answer the research questions (\cref{sec:method_jedi}) and validate our framework empirically and qualitatively (\cref{sec:results}, \cref{sec:qualitative_analysis}).
\end{enumerate}
\section{Performance Asymmetry}
\label{sec:performance_bias}
\begin{figure}[h]
    \centering
    \large
    \includegraphics[width=0.48\textwidth]{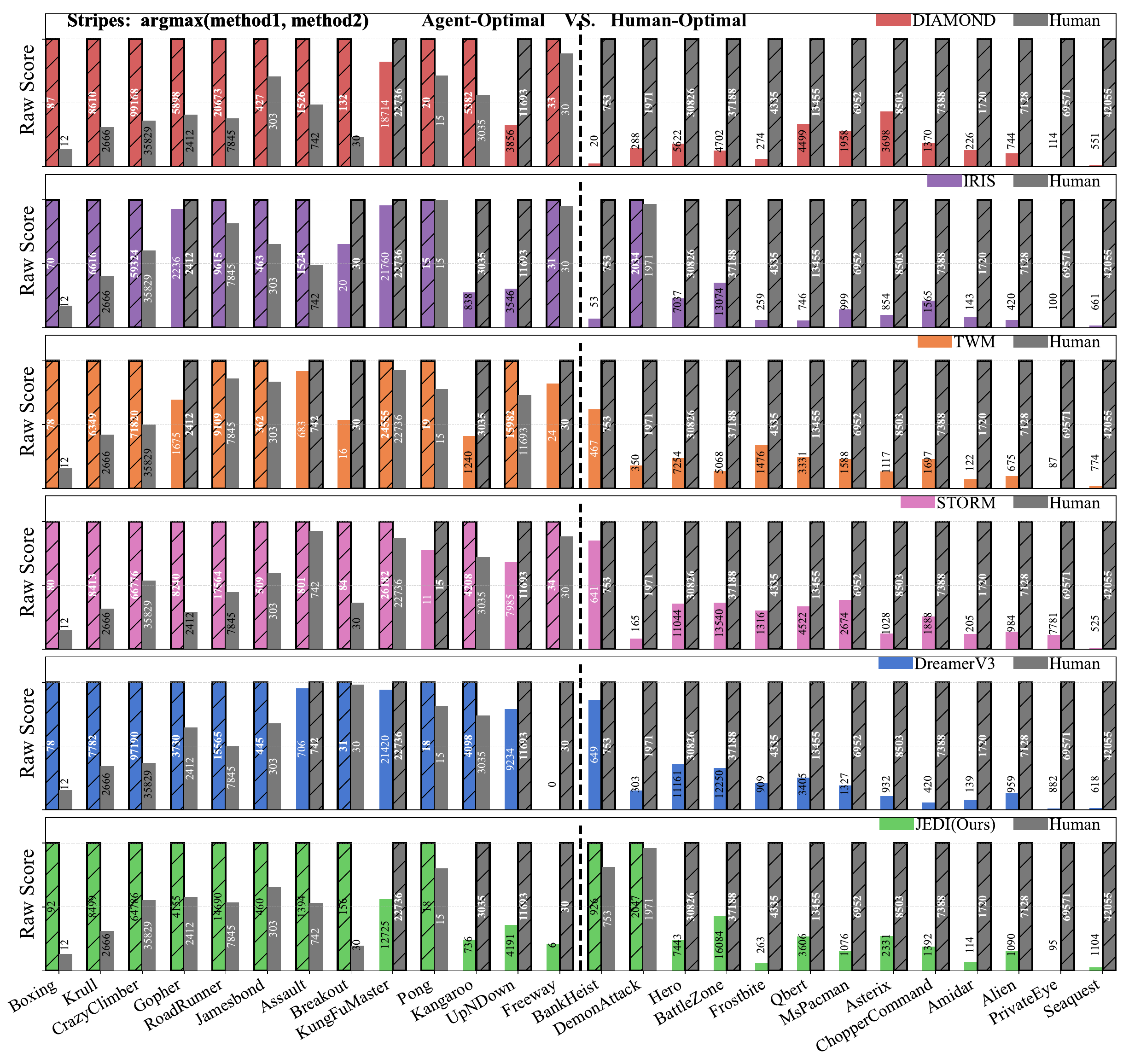}
    \caption{Performance asymmetry in MBRL algorithms. Bar heights are normalized by the max over the method and human score for every game. Bold refers to the argmax over the agent and human score. MBRL is critically failing on certain tasks while performing exceedingly well in others.}
     % This is especially pronounced in pixel-based agents like DIAMOND and IRIS. JEDI mitigates this performance asymmetry with better performance in the Human-Optimal set. 
    \label{fig:mbrl_vs_human_rawscore}
\end{figure}
Pre-existing methods that achieved a overall mean HNS of around $1.0$ did so with super-human performance in only $8-11$ out of the $26$ Atari100k tasks (\cref{tab:atari100k_FULL_NUMBERS}). Since HNS is $\frac{\textnormal{Agent Score}-  \textnormal{Random Score}}{\textnormal{Human Score} - \textnormal{Random Score}}$, this implies that exceptional performance in a minority of tasks with low human scores is overcompensating for sub-human performances in majority of tasks. Moreover, none of the baselines outperform the human score in  12 out of 13 Human-Optimal tasks. Indeed, \cref{fig:mbrl_vs_human_rawscore} reveals a drastic skew across task performance.

  % \citet{alonso2024diamond} was only able to achieve SOTA performance in only 10 tasks out of 26 tasks and superhuman performance on 11 out of 26 tasks  ; despite this, it is able to achieve the SOTA mean HNS of 1.40
\begin{table} [h] % 'r' means right side, width approx 55% text width
\scriptsize
\centering
\caption{Agent-Optimal and Human-Optimal Atari100k tasks with HNS.}
\label{tab:optimal_tasks}
% \resizebox{0.9\linewidth}{!}{%
\begin{tabular}{llll}
\toprule
\textbf{Agent-Optimal Tasks}          & HNS          & \textbf{Human-Optimal Tasks}          & HNS         \\
\midrule
\textit{Boxing}              & 6.35         & \textit{BankHeist}           & 0.59        \\
\textit{Krull}               & 5.33         & \textit{DemonAttack}         & 0.31        \\
\textit{CrazyClimber}        & 2.52         & \textit{Hero}                & 0.27        \\
\textit{Gopher}              & 1.72         & \textit{BattleZone}          & 0.25        \\
\textit{RoadRunner}          & 1.65         & \textit{Frostbite}           & 0.22        \\
\textit{Jamesbond}           & 1.52         & \textit{Qbert}               & 0.21        \\
\textit{Assault}             & 1.36         & \textit{MsPacman}            & 0.20        \\
\textit{Breakout}            & 1.25         & \textit{Asterix}             & 0.093       \\
\textit{KungFuMaster}        & 1.03         & \textit{ChopperCommand}      & 0.088       \\
\textit{Pong}                & 1.03         & \textit{Amidar}              & 0.085       \\
\textit{Kangaroo}            & 0.85         & \textit{Alien}               & 0.077       \\
\textit{UpNDown}             & 0.78         & \textit{PrivateEye}          & 0.032       \\
\textit{Freeway}             & 0.75         & \textit{Seaquest}            & 0.014       \\
\bottomrule
\end{tabular}
% }
% \vspace{-2mm}
\end{table}
In order to study Performance Asymmetry, we take inspiration from fairness research to identify a \textbf{Protected Attribute} \citep{barocas2023FairnessTB} for every task so we can observe the problem clearly. To achieve this,  we averaged the mean HNS of four pre-existing baselines that were among the first to approach or surpass the mean HNS of $1$ on Atari100k: IRIS \citep{micheli2022iris}, TWM \citep{robine2023TWM}, DreamerV3 \citep{hafner2023Dreamerv3}, and STORM \citep{zhang2023storm}. Concretely, we define tasks on which the averaged agent's mean HNS exceeds 75\% as \textbf{\textit{Agent-Optimal}} and those falling below as \textbf{\textit{Human-Optimal}}. This classification provides a balanced set of 13 tasks per subset (\cref{tab:optimal_tasks}).
\subsection{Existing Evaluation Metrics}
\label{sec:existing_evaluation}
 The core issue with existing metrics is due to (i) the arithmetic mean being invariant to extreme values and (ii) the large value range of HNS\footnote{HNS Range for existing methods is $[-0.08, 8.3]$} resulting, in large values dominating over low values. As a motivating example, with arbitrary metrics on a benchmark with two tasks comparing methods A and B :
 \begin{enumerate}[noitemsep, topsep=0pt, label=(\roman*)]
    \item A: $(1,10)$, B: $(5.5,5.5)$; Arithmetic mean is equal, masking the extreme scores in A.
    \item A: $(0.1, 11)$, B: $(0.2, 10)$; Arithmetic mean of A ($5.55$) outperforms B ($5.05$). However, B outperforms A in the first task by 100\% and yet, A only outperforms B on the second task by 10\%. 
 \end{enumerate}
 % which addresses the problem of statistical uncertainty in RL by recommending stratified bootstrap confidence intervals, Interquartile Mean (IQM), and optimality gap.
 % However, the use of arithmetic mean for aggregation across tasks, leaves Performance Asymmetry hidden due to extreme values dominating over others
 This issue is prevalent in current arithmetic mean aggregates, such as stratified bootstrap confidence intervals, Inter Quartile Mean (IQM), and the optimality gap proposed by \citet{agarwal2021precipice}. This can be seen from \cref{fig:jedi32_vs_baselines_full} where DIAMOND \citep{alonso2024diamond} achieved state-of-the-art (SOTA) for almost all metrics with reasonable error bars, yet the Performance Asymmetry in \cref{fig:T_plot_all_results} and \ref{fig:mbrl_vs_human_rawscore} is not visible at all. Moreover, there are other issues: IQM eliminates entire tasks rather than statistical anomalies, and the optimality gap is blind to progress beyond HNS of $1.0$ (\cref{sec:discussion_other_aggregates}).
 
\citet{agarwal2021precipice} also proposed performance profiles to address the disparity in task performance. However, the typical plot and large range of HNS draw focus towards the high HNS regime, concealing the low HNS regime (\cref{fig:performance_profiles}). 
% Issue with standard deviation. 
% - emphasise that we want a metric that celebrates wins ACROSS the board, while highlighting balance. Dont want metrics that if optimized for, can lead to adverse outcomes in agent performance on some games e.g. use STD
\subsection{Sym-HNS}
\label{sec:sym-hns_evaluation}
We strive for an aggregate that celebrates progress on all tasks in a balanced manner, exposing extreme data spread while preventing the delta of large values from dominating. We introduce:
\begin{align}
    \textnormal{Sym-HNS} = \frac{2\mu_{\tau}(\textnormal{AO)}\cdot\mu_{\tau}(\textnormal{HO)}}{\mu_{\tau}(\textnormal{AO)}+ \mu_{\tau}(\textnormal{HO)}}
\end{align}
\begin{wrapfigure}{r}{0.18\textwidth}
    \centering
    \includegraphics[width=0.18\textwidth]{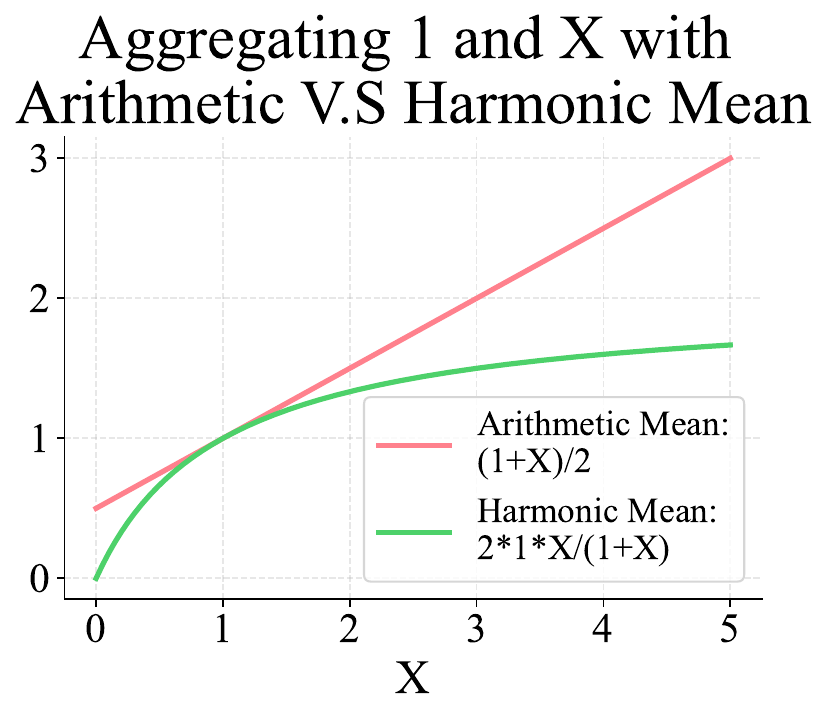}
    \caption{Harmonic v.s. Arithmetic Mean }
    \label{fig:harmonic_vs_arithmetic_mean}
\end{wrapfigure}
where $\mu_{\tau}$(AO/HO) refers to the mean HNS on the Agent/Human-Optimal subset respectively. Sym-HNS leverages the harmonic mean, as inspired by the F1-Score \citep{F1vanRijsbergen1979}, effectively balancing two equally important metrics by discouraging extreme and dominant values (\cref{fig:harmonic_vs_arithmetic_mean}). Regarding the example in \cref{sec:existing_evaluation}
 \begin{enumerate}[noitemsep, topsep=0pt, label=(\roman*)]
    \item A: $(1,10)$, B: $(5.5,5.5)$; Sym-HNS is 1.82 and 5.5, revealing the spread in A.
    \item A: $(0.1, 11)$, B: $(0.2, 10)$; Sym-HNS is 0.2 and 0.39, rewarding the disproportionate gain by B in the first task  
 \end{enumerate}
Most importantly, Sym-HNS clearly reflects the Performance Asymmetry exhibited by each method\footnote{\cref{sec:discussion_other_aggregates} discusses our justifications against other aggregates} (\cref{fig:T_plot_all_results}).
% \begin{enumerate}[noitemsep, topsep=0pt, label=(\roman*)]
%     \item A: $[1,10]$, B: $[5.5,5.5]$ Sym-HNS(A)=$1.82$, Sym-HNS(B)=$5.5$
%     \item A: $[0.1, 11]$, B: $[0.2, 10]$, Sym-HNS(A)=$0.20$, Sym-HNS(B)=$0.39$
% \end{enumerate}
% \subsection{Discussion} 
% \label{sec:asym_discussion}
% All of the pre-existing methods have their super-human performances within the Agent-Optimal tasks (\cref{tab:atari100k_FULL_NUMBERS}), except for IRIS on \textit{DemonAttack}. Crucially, this means \textbf{none of the baselines outperform the human score in  12 out of 13 Human-Optimal tasks}. It is clear that existing MBRL agents overfits to the Agent-Optimal tasks. This presents a major research opportunity; 
% \subsection{Performance Asymmetry Analysis}
% \label{sec:pronounced_asymmetry_hypothesis}
% This work focuses on reversing the worsening trend of Performance Asymmetry over time, specifically on the latest SOTA agent, DIAMOND, which has the most asymmetry (Figure \ref{fig:T_plot_all_results}). 
% The median number of actions in the Agent-Optimal tasks is 9 as compared to 18 in Human-Optimal tasks.
% There are 7 Shooter tasks in the Human-Optimal tasks and only 2 Shooter tasks in the Agent-Optimal tasks.
\subsection{Agent-Optimal V.S Human-Optimal Tasks}
\label{sec:agent_vs_human_task_analysis}
\begin{wrapfigure}{r}{0.20\textwidth}
    \centering
    \includegraphics[width=0.20\textwidth]{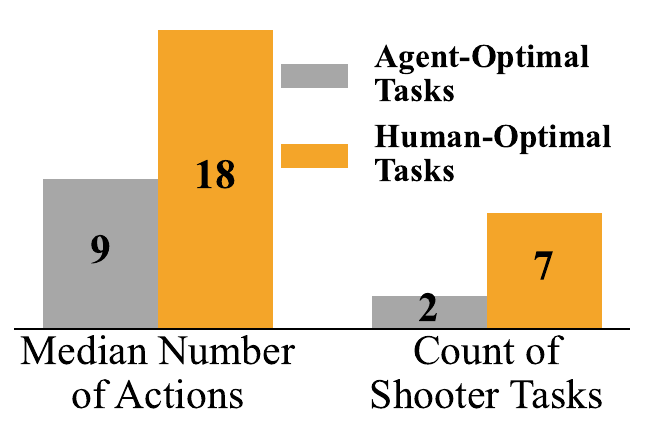}
    \caption{Quantitative features of Agent/Human-Optimal tasks (\cref{tab:tasks_with_num_actions_shooter_task}). }
    \label{fig:action_space_shooter_comparison}
\end{wrapfigure}

We begin our analysis by investigating the quantitative differences between the two subsets, revealing two distinctions: action space and the number of Shooter tasks. We define Shooter tasks as tasks where the agent has a \textit{FIRE} action that activates a ranged weapon to neutralize or deter enemies. Human-Optimal tasks have a much higher action space and number of Shooter tasks as shown in \cref{fig:action_space_shooter_comparison}.
% Human-Optimal tasks are more complex with higher action space and number of shooter tasks.
% as they necessitate the correct timing of the \textit{FIRE} action to neutralize or deter an enemy a few time steps into the future. As such

This suggests that Human-Optimal tasks are more complex in general. Firstly, with every additional action, the search space increases exponentially. Secondly, Shooter tasks are much more complex compared to Non-Shooter tasks. Returns of states in Shooter tasks have very high variance as the same state can have drastically different outcomes depending on timing of \textit{FIRE} and future actions thereafter. For example, in \textit{BankHeist}, the \textit{FIRE} action releases a dynamite that explodes a few seconds later; successfully neutralizing an enemy versus walking into your own demise leads to very different returns (\cref{fig:qualitative_analysis}). 
% With the identification of Agent/Human-Optimality of each task, we can clearly observe Performance Asymmetry by plotting the overall performance of pre-existing MBRL agents on the two sets of tasks as shown in Fig \ref{fig:T_plot_all_results}. While, this discovery sparks fundamental questions into the overall performance of  MBRL algorithms, it also opens up future research into identify the root causes of poor performance in Human-Optimal tasks, enabling meaningful research discoveries.    
% Due to the worrying trend of increasing Performance Asymmetry, specifically with the latest SOTA algorithm, DIAMOND \citep{alonso2024diamond} exhibiting the most asymmetry,  in this work.
% \subsubsection{Agent-Optimal versus Human-Optimal Tasks: Task Complexity}
% Action space and shooter tasks.
% We propose that MBRL agents perform poorly on Human-Optimal tasks because they are much more complex than Agent-Optimal tasks. Furthermore, the exceptionally poor performance on Human-Optimal tasks by pixel-based agents as compared to latent-based agents is due to the curse of dimensionality \citep{MORE DREAMER GUYS}. This is supported by the fact that the main difference between agents with high Performance Asymmetry and those with less is the use of pixel versus latents (\cref{fig:T_plot_all_results}). 
\subsection{Curse of Dimensionality: Underperformance of Pixel-Based Methods on Human-Optimal Tasks}
Our first insight on algorithms hinges on the stark finding that pixel-based agents perform much worse in Human-Optimal tasks compared to latent-based agents (\cref{fig:T_plot_all_results}). This suggests that latent world models can help with Human-Optimal performance.

\begin{wrapfigure}{r}{0.25\textwidth}
    \centering
    \includegraphics[width=0.25\textwidth]{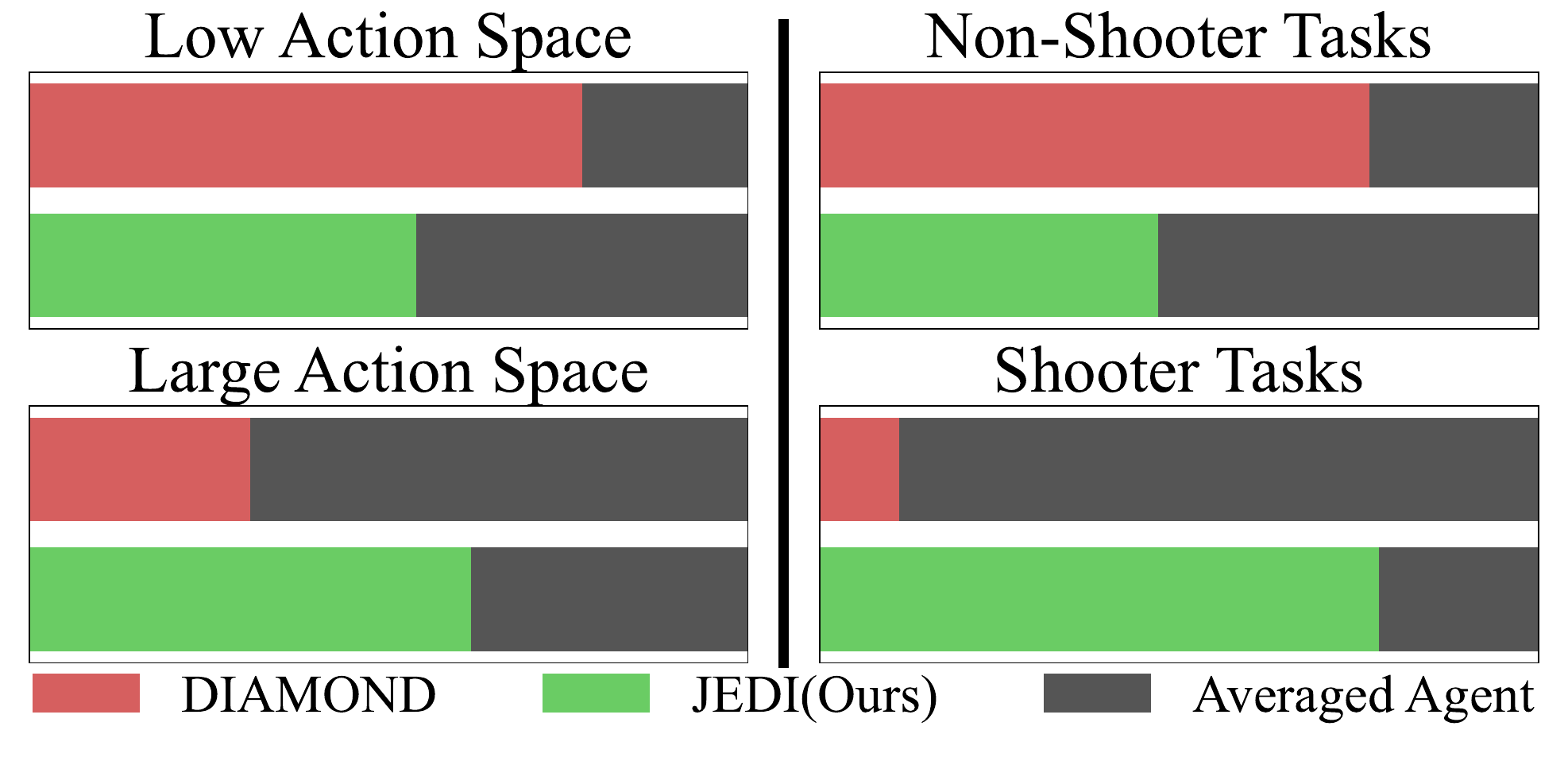}
    \caption{Mean HNS of DIAMOND and JEDI against the averaged agent \cref{tab:optimal_tasks}. DIAMOND underperforms in high Action Space and Shooter tasks whereas JEDI does not.}
    \label{fig:diamond_jedi_versus_others_actionspace_shooter}
\end{wrapfigure}

\textbf{Action Space Complexity}. We hypothesize that the pixel-based agents perform worse in Human-Optimal tasks due to poorer training efficiency \citep{watter2015embed_Latent, oh2017value_Latent, gregor2019shaping_Latent, Hafner2020Dreamer}. This is because their RL network input is in $\mathbb{R}^{64 \times 64 \times 3}$. The exponential increase in search space due to higher action space disproportionately affects pixel-based agents compared to latent-based agents. Moreover, given the same number of environment steps, pixel-based RL networks need to perform both representation learning and RL training, whereas the RL networks of latent-based agents only have to perform RL training. This is supported empirically in \cref{fig:diamond_jedi_versus_others_actionspace_shooter} where SOTA pixel-based agent performs significantly worse at tasks with high action space, contributing to its underperformance in Human-Optimal tasks.

\textbf{Shooter Tasks}.  
% The state where the agent releases the dynamite has a very high variance value as moving away and neutralizing the enemy will lead to high value, while walking towards the dynamite resulting in loss of a life will lead to low value (\cref{fig:qualitative_analysis}). 
% challenging task of associating similar states to the correct actions and high variance values.
Similarly, Shooter tasks are substantially harder than Non-Shooter tasks. 
It is very difficult for the actor to assign the correct probabilities to the \textit{FIRE} action, as well as for the critic to assign the correct value to states due to the high variance in outcomes. Again, this is much easier with a learned compressed latent state as opposed to an image input. This is supported empirically in \cref{fig:diamond_jedi_versus_others_actionspace_shooter} as the SOTA pixel-based agent underperforms in Shooter tasks, contributing to its poor performance in Human-Optimal tasks. 
% \begin{enumerate}
%     \item GENERAL ASYMMETRY IS BECAUSE HUMAN-OTPIMAL TASKS AND AGENT-OPTIMAL TASKS ARE FUNDAMENTALLY DIFFERENT BECAUSE OF COMPLEXITY (ACTION SPACE AND SHOOTER TASK). Thats why in general MBRL 1 order magnitude performance gap
%     \item  ANALYZE PA FIRST THEN MOTIVATE
%     \item Using this division to justify my method
%     \item first point is latent outperforms pixel in HO tasks. why? latent is more efficient in learning (cite the dreamer works), curse of dimensionality. every action space is exponential in state, action search space. with shooter task, similar states have very high variance,
%     \item comparing DIAMOND - IRIS, diamond is better, diffusion is better. discretization 
% \end{enumerate}
% \subsection{Curse of Dimensionality of Pixel-based Agents}
% we are able to identify which MBRL agents have high Performance Asymmetry.
% Otherwise, they are similar in their iterative training framework, RL algorithm of choice etc.
\begin{figure*}[!h]
    \centering
    \begin{subfigure}{0.25\textwidth}
        \centering
        \includegraphics[width=\textwidth]{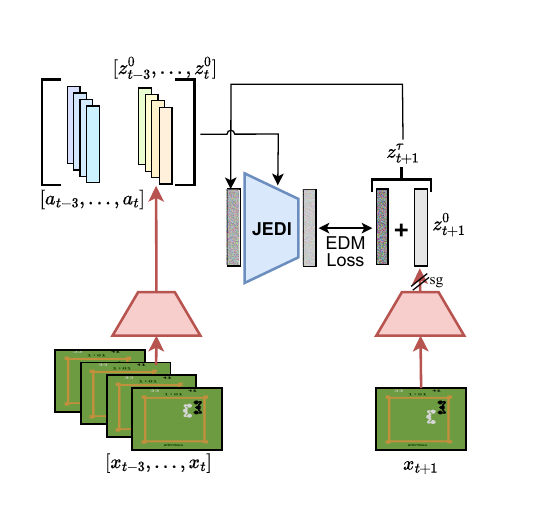}
        \caption{JEDI Training}
        \label{fig:JEDI_train}
    \end{subfigure}
    % Vertical line
    \hspace{0.1\textwidth}
    \vrule width 0.5pt
    \hspace{0.1\textwidth}
    % \hfill
    \begin{subfigure}{0.25\textwidth}
        \centering
        \includegraphics[width=\textwidth]{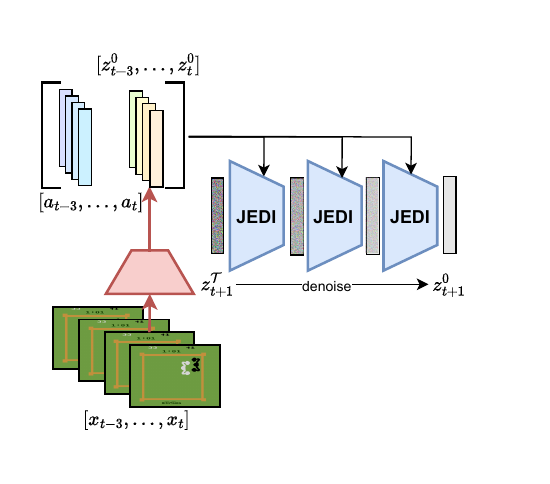}
        \caption{JEDI Inference}
        \label{fig:JEDI_inf}
    \end{subfigure}
    \caption{Joint Embedding DIffusion (JEDI) World Model. (a) During training, the images are passed through the encoder to derive the compressed latents, $z_t^0$. The noised next state $z_{t+1}^\tau$ is derived by summing the clean latent with a noise that is scaled according to sampled diffusion time step $\tau$. The diffusion model is conditioned on these inputs to predict the direction towards the next state, training both the encoder and diffusion model jointly using EDM loss \citep{karras2022EDM}. (b) During inference, the same conditioning is derived and passed into the diffusion model except that $z_{t+1}^\tau$ is replaced with noise, and the model iteratively denoise to arrive at the clean next state.}
    \label{fig:jedi_training_inference}
\end{figure*}

\subsection{Diffusion for Better Agent-Optimal Performance}
\label{sec:visual_tasks_analysis}
We observe that out of the two pixel-based agents, DIAMOND significantly outperforms IRIS in Agent-Optimal tasks (\cref{fig:T_plot_all_results}). This suggests that using a diffusion dynamics model can help with Agent-Optimal performance, as it is the main difference between the two methods. In the paper, \citet{alonso2024diamond} credited their outperformance over IRIS to the diffusion models' high fidelity generation, enabling better performance in visual tasks, most notably \textit{Breakout}, an Agent-Optimal task. Indeed, DIAMOND overwhelmingly outperforms in \textit{Breakout}, with a HNS of 4.54 versus the averaged agents' HNS of 1.25 (\cref{tab:optimal_tasks}), contributing significantly to its outperformance on Agent-Optimal tasks. 

% We analyzed that the most visual tasks are Breakout and Assault by comparing the results of pixel-based agents against the best performing scores of latent-based agents. 
% TAKE THE DELTA BETWEEN DIAMODN AND IRIS TO MOTIVATE DIFFUSION
% The main claim of DIAMOND is that the use of diffusion models, which excel in generating high fidelity images, enables outperformance in high visual detail tasks---with the most notable being \textit{Breakout}\footnote{\textit{Breakout} is referenced in multiple figures in DIAMOND for qualitative generation}, which happens to be an Agent-Optimal task. Indeed, DIAMOND overwhelmingly outperform the other baselines in \textit{Breakout}, with a HNS of 4.54 versus the averaged agents' HNS of 1.25 (\cref{tab:optimal_tasks}), contributing significantly to the pronounced Performance Asymmetry. 
To this end, we propose the following research questions: 
\begin{enumerate}[noitemsep, topsep=0pt]
    \item Can the striking Performance Asymmetry of a pixel-based agent be mitigated with a latent space world model?
    \item Can the use of diffusion in a latent dynamics model overcome the underperformance of latent-based agents on high visual tasks?
\end{enumerate}

% \begin{table*}[t]
% \caption{\textbf{Claim$\rightarrow$Evidence map} for \emph{HypoSpace}.}
% \label{tab:claim_evidence_hypospace}
% \vspace{-0.4em}
% \centering
% \scriptsize
% \setlength{\tabcolsep}{3pt}
% \renewcommand{\arraystretch}{1.05}
% \begin{tabular}{p{0.07\linewidth} p{0.47\linewidth} p{0.33\linewidth} p{0.10\linewidth}}
% \hline
% \textbf{ID} & \textbf{Claim} & \textbf{Evidence} & \textbf{Metric}\\
% \hline
% C1 & Underdetermination needs \textbf{set coverage}, not 1-shot accuracy. &
% Define admissible set $H_O$; treat LLM as sampler. &
% RR \\

% C2 & VR/NR/RR separate validity, novelty, coverage. &
% Formal defs + canonicalizers. &
% VR/NR/RR \\

% C3 & Theory: peaked $p(h)$ implies slow recovery; tails hard. &
% Expected coverage bound. &
% RR vs $|H_O|$ \\

% C4 & HypoSpace: 3 tasks with exact validators, controllable $|H_O|$. &
% DAG / Gravity / GP constructions. &
% Sound/ctrl \\

% C5 & Empirics: VR stays high; NR/RR collapse as $|H_O|$ grows. &
% Tables/curves across difficulty. &
% RR, dup-rate \\

% C6 & Simple fix improves coverage (reduces simplicity bias). &
% Complexity-stratified decoding. &
% RR$\uparrow$ \\
% \hline
% \end{tabular}
% \vspace{-0.6em}
% \end{table*}
\begin{table*}[t]
\caption{\textbf{Claim$\rightarrow$Evidence map} for \emph{Performance Asymmetry in Model-Based RL / JEDI}.}
\label{tab:claim_evidence_jedi}
\vspace{-0.4em}
\centering
\tiny
\setlength{\tabcolsep}{3pt}
\renewcommand{\arraystretch}{1}
\begin{tabular}{p{0.01\linewidth} p{0.41\linewidth} p{0.45\linewidth} p{0.07\linewidth}}
\hline
\textbf{ID} & \textbf{Claim} & \textbf{Evidence} & \textbf{Metric}\\
\hline
C1 & Conventional aggregates masks \textbf{Performance Asymmetry}: agents excel on Agent-Optimal(AO) but fail on Human-Optimal(HO) tasks. &
Fig.\ref{fig:T_plot_all_results}: large gap; SOTA pixel agent worst on HO despite best overall mean HNS; 21$\times$ HO/AO disparity. &
HO/AO mean HNS gap \\
% define HO/AO using performance by seminal works
\midrule
C2 & A more balanced evaluation can be achieved by (i) HO/AO partition and (ii) harmonic mean&
Fig\ref{fig:jedi32_vs_baselines_full}: Performance Asymmetry invisible in conventional aggregates; Sec.\ref{sec:performance_bias}-\ref{sec:sym-hns_evaluation}: Sym-HNS uses harmonic mean over HO/AO to address issues with conventional arithmetic mean & Aggregates;
Sym-HNS \\
\midrule
C3 & Pixel agents underperform on HO due to curse of dimensionality + $\uparrow$HO task complexity.&
Fig.\ref{fig:action_space_shooter_comparison}: HO has $\uparrow$ action space + $\uparrow$ shooter tasks; Fig.\ref{fig:T_plot_all_results},\ref{fig:diamond_jedi_versus_others_actionspace_shooter}: pixel agents performs worse on these tasks; motivates latent world model&
mean HNS \\
\midrule
C4 & SOTA pixel agent outperform on AO due to diffusion and visual tasks in AO.&
Fig.\ref{fig:T_plot_all_results}: SOTA pixel-agent with diffusion outperforms prior pixel-agent in AO, and it outperforms baselines (4.25v.s.1.25) on \textit{Breakout}(AO)$\therefore$$\uparrow$AO HNS; motivates diffusion & mean HNS \\
\midrule
C5 & \textbf{JEDI}: novel end-to-end \textbf{latent diffusion} world model with JEPA-style training (no reconstruction/distillation/network hidden activations)&
Fig.\ref{fig:jedi_training_inference} + Eq.(\ref{eq:world_model_components}-\ref{eqn:jedi_rew_loss}): latent EDM loss trains encoder+dynamics end-to-end with JEPA; Sec.\ref{related_works}: first to show that off-the-shelf diffusion is directly compatible with JEPA &
- \\
\midrule
C6 & JEDI \textbf{reverses asymmetry trend} and achieves SOTA on \textbf{Sym-HNS} and HO while staying competitive on AO and full Atari100k. &
Fig.\ref{fig:T_plot_all_results}: best Sym-HNS; Fig.\ref{fig:jedi32_vs_baselines_HO}: SOTA on HO aggregates; Fig.\ref{fig:jedi32_vs_baselines_AO}-\ref{fig:jedi32_vs_baselines_full}: competitive performance&
Sym-HNS; Aggregates \\
\midrule
C7 & JEDI performs well on visually bottlenecked tasks & Sec.\ref{sec:results}: definition; Tab.\ref{tab:visual_bottleneck}: achieves SOTA in \textit{Breakout}, runner-up in \textit{Assault} &
HNS \\
\midrule
C8 & JEDI is \textbf{efficient} (latent compression) and robust in stochastic settings; ablations show JEDI gradients are effective (esp. HO). &
Fig.\ref{fig:compute_efficiency}: 43\% lower memory + 3$\times$sampling speed; Fig.\ref{fig:compute_and_runtime_comparison}: 2$\times$training speed; Fig.\ref{fig:stoch_experiments}: random frameskip wins; Fig.\ref{fig:new_ablation_study}: altering JEDI-grad hurts HO (hurts AO less). &
Mem/speed; mean HNS $\Delta$ \\
\hline
\end{tabular}
\vspace{-0.6em}
\end{table*}
% , learning encoder from diffusion denoising gradients
\section{JEDI: Joint Embedding DIffusion World Model World Model}
\label{sec:method_jedi}
% The JEDI world model learns a latent space encoder in an end-to-end fashion, while using a diffusion model as the dynamics model.
To answer the above research questions, we introduce the Joint Embedding DIffusion World Model (JEDI), which learns an end-to-end latent space encoder while leveraging a diffusion dynamics model (\cref{fig:jedi_training_inference}). Mathematically:
\begin{equation}
\textbf{World Model}
    \begin{cases}
        \text{Encoder:} \\ z_t^0 = \mathbf{E}_{\phi}(x_t), \\
        
        \text{Latent Diffusion Dynamics Model:} \\ \hat{z}_{t+1}^0 \sim  \mathbf{S}(\mathbf{D}_{\theta}(\hat{z}_{t+1}^\tau,Z_t^\tau)),\\
        
        \text{Reward \& Termination}  \\ (\hat{r}_t, \hat{d}_t) = \mathbf{R}_{\psi}(z_t^0),
    \end{cases}
\label{eq:world_model_components}
\end{equation}
% $\textnormal{LN}$ refers to the logit-normal distribution \citep{karras2022EDM}{}
where $t$ is the environment time step, $x_t$ is the environment observation, $\mathbf{E}_{\phi}$ is the encoder, $z^0_t$ refers to the clean latent at environment time step $t$, $x_t \in [0,1]^{64\times64\times3}$ is the environment image input and $z_t \in [-3,3]^{16\times8\times8}$ is the latent state, $\tau$ is the diffusion time step, drawn from the Log-Normal (LN) distribution, determining the noise schedule $\sigma(\tau)$ with $0$ being clean latent state, $\mathbf{D}_{\theta}$ is the single-step conditioned denoising operator and $\mathbf{S}$ is the ODE or SDE solver for multi-step denoising inference. $r_t\in \{-1,0,1\}$ is the reward and $d_t\in\{0,1\}$ is the episode termination flag. We follow the pre-conditioning diffusion paradigm as introduced in \citep{karras2022EDM}: 
\begin{equation}
\mathbf{D}_{\theta}(z_{t+1}^\tau,\,Z_{t}^\tau)=c_{\mathrm{skip}}^\tau\,z_{t+1}^\tau+c_{\mathrm{out}}^\tau\,\mathbf{F}_{\theta}\bigl(c_{\mathrm{in}}^\tau\,z_{t+1}^\tau,\;Z_{t}^\tau\bigr)\,
\end{equation}
where, $c_{\mathrm{noise}}^\tau$ is a fixed transformation of $\tau$, $Z_t^\tau := (c_{\mathrm{noise}}^\tau, z^0_{t-3:t},a_{t-3:t})$, and $c_{\mathrm{skip}}^\tau, c_{\mathrm{out}}^\tau, c_{\mathrm{in}}^\tau$ are preconditioning variables depending on sampled noise for better behavior of $\mathbf{F}_\theta$, the neural network parameterized by $\theta$. 

% $c_{\mathrm{skip}}^\tau , z_{t+1}^\tau$ passes part of the noisy input directly to the output, improving gradient flow

\textbf{Joint Embedding DIffusion loss}. The loss function of the latent diffusion dynamics model, $\mathcal{L}_{dyn}(\theta, \phi)$ is re-parameterized to:
{\setlength{\abovedisplayskip}{4pt}%
 \setlength{\belowdisplayskip}{6pt}%
 \setlength{\abovedisplayshortskip}{0pt}%
 \setlength{\belowdisplayshortskip}{0pt}%
\begin{equation}
\begin{split}
\E_{z_{1:T} \sim q, x_{1:T}\sim p, \tau \sim \textnormal{LN}}\Biggl[
\Biggl\| \sum_{t=1}^{T}
\mathbf{F}_{\theta}\bigl(c_{\mathrm{in}}^\tau \mathrm{sg}(z_{t+1}^\tau),Z_{t}^\tau\bigr)
\\ - \frac{1}{c_{\mathrm{out}}^\tau}\bigl(\mathrm{sg}(z_{t+1}^0) - c_{\mathrm{skip}}^\tau\, \mathrm{sg}(z_{t+1}^\tau) \bigr) 
\Biggr\|^2 \Biggr]  
\end{split}
\label{eqn:jedi_dyn_loss}
\end{equation}
}
where $q(z_t) = (\mathbf{E}_\phi)_{\#}\,p(x_t)$ is the deterministic push forward mapping of $\mathbf{E_\phi}$ on the randomly sampled $x_t$ and $\mathrm{sg}$ refers to stop-grad. 
% Sampling $\tau$ from a logit-normal prior \textnormal{LN} ensures the middle noise level are visited more often during training.

During JEDI training (\cref{fig:JEDI_train}), the encoder $\mathbf{E}_\phi$ first derives $z_{t-3:t+1}^0$ from $x_{t-3:t+1}$. $z_{t+1}^\tau$ is derived by adding $z_{t+1}^0$ with stop-grad to the sampled noise which is scaled according to $\tau$. It is then passed into $\mathbf{F_\theta}$ together with $Z_t^\tau$ and it regresses towards the direction of the next latent state using \cref{eqn:jedi_dyn_loss}. During inference (\cref{fig:JEDI_inf}), the conditioning is derived in the same manner, except $z_{t+1}^\tau$ is simply random noise to start the iterative denoising to arrive at the clean latent state (\cref{alg:jedi}).

The Cross-Entropy (CE) loss for the Reward and Termination model, $\mathcal{L}_{r\&d}(\psi,\phi)$, is:
% \begin{equation}
% \begin{split}
% \E_{z_{1:T}^0 \sim q, (x_{1:T},r_{1:T},d_{1:T}) \sim p}
% \sum_{t=1}^T \textnormal{CE}(\; \mathbf{R_\psi}(z_t^0), (r_t,d_t) \;)
% \label{eqn:jedi_rew_loss}
% \end{split}
% \end{equation}
{\setlength{\abovedisplayskip}{4pt}%
 \setlength{\belowdisplayskip}{4pt}%
 \setlength{\abovedisplayshortskip}{0pt}%
 \setlength{\belowdisplayshortskip}{0pt}%
\begin{equation}
\begin{split}
\E_{z_{1:T}^0 \sim q, (x_{1:T},r_{1:T},d_{1:T}) \sim p}
\sum_{t=1}^T \textnormal{CE}(\; \mathbf{R_\psi}(z_t^0), (r_t,d_t) \;)
\label{eqn:jedi_rew_loss}
\end{split}
\end{equation}
}
\textbf{Preventing Representation Collapse}. The encoder learns from both \ref{eqn:jedi_dyn_loss} and \ref{eqn:jedi_rew_loss} where the gradients flow back to $\mathbf{E_\phi}$ through $z_{t}^0$. In order to prevent representation collapse, stop-grad is applied to $z_{t+1}^0$ and the learning rate of $\mathbf{E}_\phi$  is 0.3 that of $\mathbf{F}_\theta$, as inspired from JEPA and TDMPC2 \citep{assran2023I-JEPA,hansen2023tdmpc2}. Crucially, this enables $\mathbf{E_\phi}$ to learn directly from $\mathbf{F}_\theta$'s denoising loss gradients without any reconstruction loss. 

\textbf{Stabilising Diffusion in Latent Space}. Conventional image diffusion has a natural clamp during denoising inference as $x_t\in[0,1]$. To ensure that the latent denoising process is stable, we impose a differentiable clamp function, $C(z_t^0) = \mathrm{tanh}(z_t^0/s)s$, scaled by a factor, $s=3$, on the outputs of $\mathbf{E_\phi}$ as well as on $\mathbf{D_\theta}$.
% We also place the single-step prediction of $\mathbf{D}_\theta$ for each time step with attached gradients as the input to the JEDI loss at the next time step. This enables the gradients from future states' denoising loss to flow back to the encoder. However, this means that $\mathbf{E_\phi}$ only sees the image observations at the starting time steps. Hence,
With every batch of trajectory sampled, we randomly switch with uniform probability between using the output of $\mathbf{D}_\theta$ and $\mathbf{E}_\phi$ for the subsequent time step's JEDI loss. This is done to balance the capturing of near-horizon information while providing sufficient input data to the encoder.

\textbf{Policy}. The policy is trained in an actor critic framework \citep{hafner2020Dreamerv2,alonso2024diamond} and the intentional use of REINFORCE \citep{williams1992REINFORCE} ensures that JEDI does not incur expensive gradient computation through the multi-step denoising of the diffusion dynamics model. 
% The actor critic networks are trained in a similar fashion to \citet{alonso2024diamond} 
% This has been effective on Atari100k \citep{hafner2020Dreamerv2}. 
% To train the LSTM-based actor-critic network~\citep{hochreiter1997LSTM}, we freeze JEDI world model and rollout latent trajectories. 
% After the policy finishes its current training iteration, its parameters alongside JEDI encoder's parameters are frozen and both are deployed in the actual environment to collect more data and the cycle continues. 

We highlight that we directly build upon the implementation of the SOTA pixel-based diffusion agent by \citet{alonso2024diamond} with minimal changes in order to attribute as much of the emergent differences in performance to JEDI (\cref{appendix:hyperparameters}).
% \section{Evaluation}
% \label{sec:fairness_evaluation}
% One potential explanation for generic Performance Asymmetry is that the machine learning community as a whole had been overfitting to Agent-Optimal tasks (i.e. tasks with naturally high HNS). This is because aggregate metrics such as mean HNS across all games is directly proportional to absolute score of each game i.e. a 2x improvement on tasks with high HNS, has an outsized as opposed to a 2x improvement on tasks with low HNS. 
% we show our results over the two sets in \cref{tab:atari100k_aggregate_statistics} and
% \vspace*{-3mm}
\section{Results and Discussion}
\label{sec:results}
% and report the mean scores achieved by the final policy on 100 test episodes
 % Our experiment runtime is  ~2 days and memory requirement is ~7GB.
% We begin our evaluation with a systematic investigation of the asymmetry present in the Atari100k benchmark through the lens of fairness research, showing that JEDI bridges the gap between human and agent performance by reducing asymmetry in performance. Then, we present our results on the Atari100k, demonstrating that JEDI compares favorably to SOTA agents while improving its  computational efficiency and scalability. Next, we validate our claim qualitatively, showing JEDI shows clever strategies of temporal reasoning compared to its pixel-based counterpart. Finally, we motivate our design choices through thorough ablation studies.
% For experiments, we compare the original average human scores on Atari100k \citep{kaiser2019Atari100kSimPLE, pohlen2018OrigAtari100kScores} and pre-existing MBRL agents' performance. We derive the general agent performance by averaging the HNS of 4 pre-existing MBRL works which hovers around human-level performance \citep{micheli2022iris, robine2023TWM, hafner2023Dreamerv3, zhang2023storm}. These works are amongst the first MBRL works to reach human-level performance.

\begin{figure}[t]
    \centering
    % \includegraphics[width=0.47\textwidth]{pdfs/jedi32_vs_baselines_1x4.pdf}
  % First subfigure
    \begin{subfigure}[t]{0.47\textwidth}
        \centering
        \includegraphics[width=\textwidth]{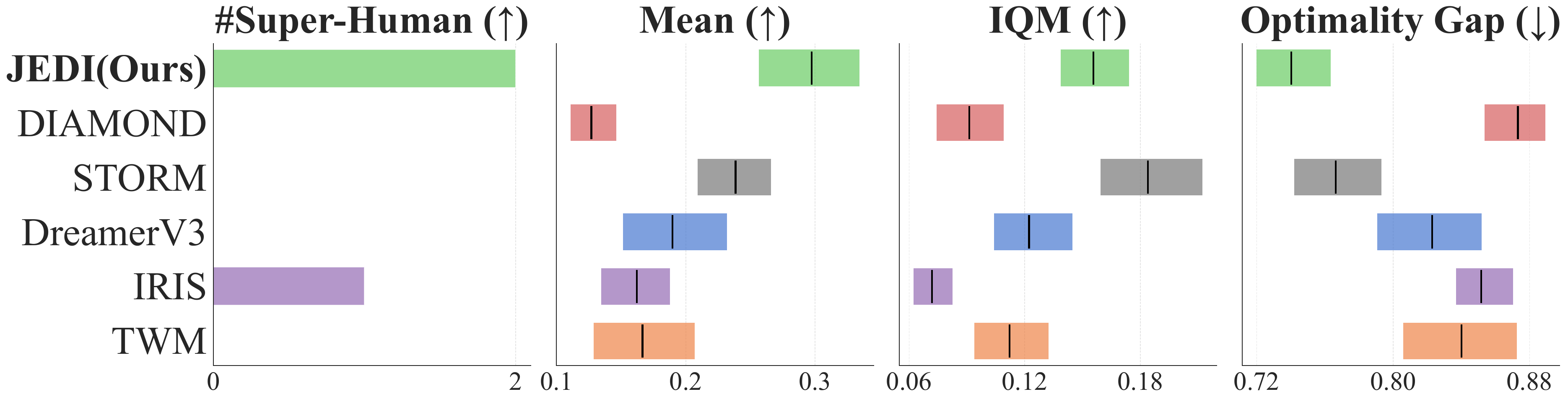}
        \caption{Aggregates on \textbf{Human-Optimal} Tasks. JEDI achieves SOTA in number of super-human tasks, mean score, optimality gap and runner-up in IQM}
        \label{fig:jedi32_vs_baselines_HO}
    \end{subfigure}
    % \hfill
    % Second subfigure
    \begin{subfigure}[t]{0.47\textwidth}
        \centering
        \includegraphics[width=\textwidth]{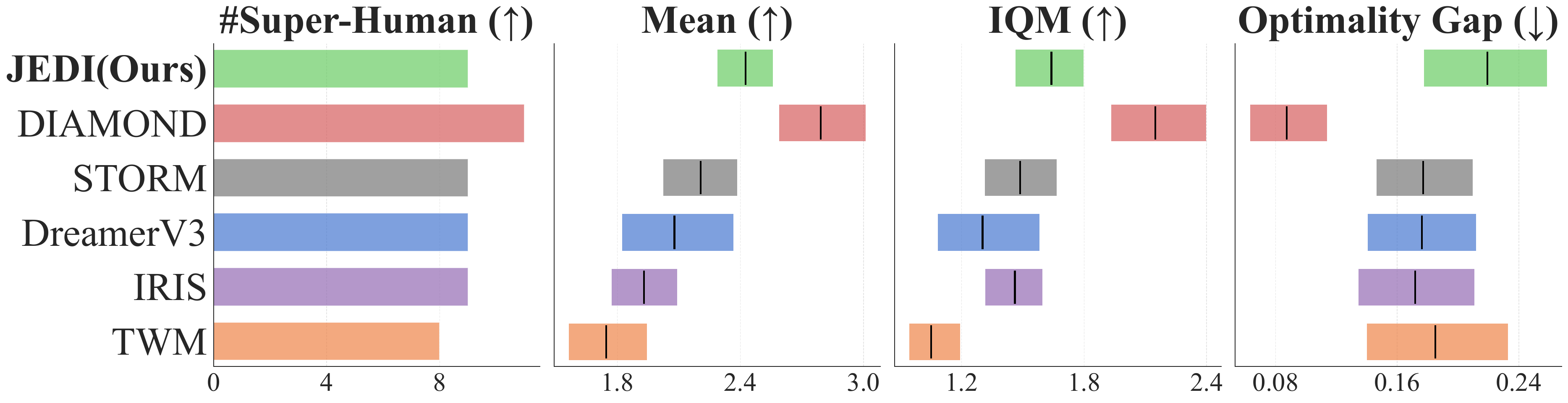}
        \caption{Aggregates on \textbf{Agent-Optimal} Tasks. JEDI achieves runner-up in number of super-human tasks, mean score, and IQM}
        \label{fig:jedi32_vs_baselines_AO}
    \end{subfigure}

    \begin{subfigure}[t]{0.47\textwidth}
        \centering
        \includegraphics[width=\textwidth]{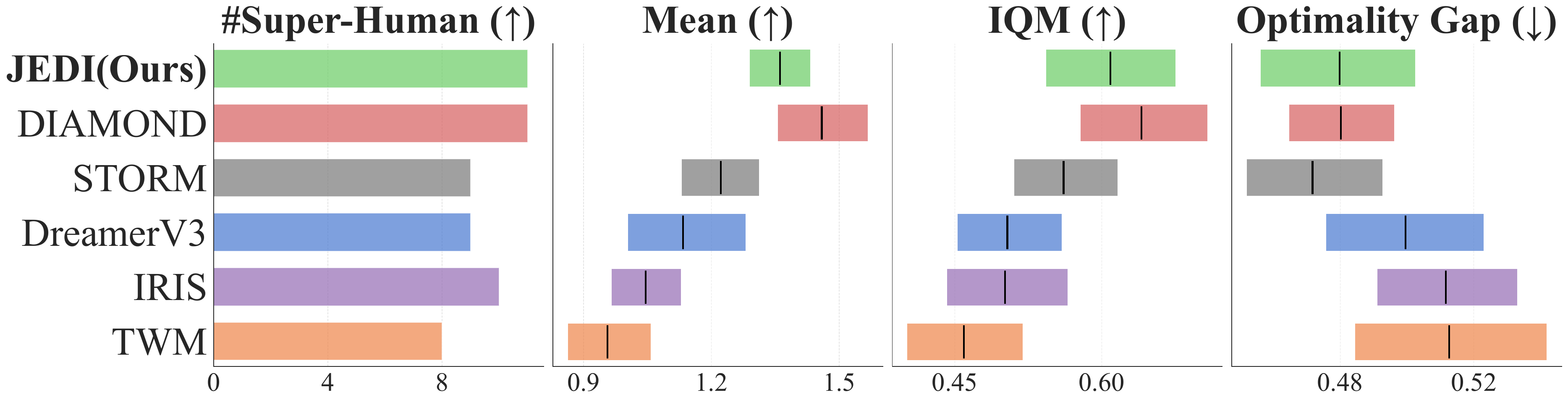}
        \caption{Aggregates on \textbf{Full Atari100k}. JEDI achieves SOTA in number of super-human tasks and runner-up in mean score, IQM and optimality gap.}
        \label{fig:jedi32_vs_baselines_full}
    \end{subfigure}
    \caption{Overall aggregates on Atari100k. Mean, Median, and Interquantile Mean (IQM) is with reference to Human Normalized Scores (↑); Optimality Gap (↓) is the average gap to HNS of 1. JEDI achieves SOTA performance in Human-Optimal tasks while maintaining competitive on Agent-Optimal tasks and the full benchmark.}
    \label{fig:jedi32_vs_baselines_HN} 
    \vspace{-4mm}
\end{figure}

\textbf{Performance Asymmetry\footnote{We validate the JEDI agent on Atari100k across 5 seeds.}.} \cref{fig:T_plot_all_results}  shows that JEDI reverses the Performance Asymmetry observed in DIAMOND, achieving SOTA performance on Sym-HNS and Human-Optimal tasks except for IQM (\cref{fig:jedi32_vs_baselines_HO})---answering our first research question. Moreover, \cref{fig:diamond_jedi_versus_others_actionspace_shooter} shows that the biased performance of DIAMOND towards low action space and Non-Shooter task is alleviated in JEDI. JEDI is also the first to achieve super-human performance on the Human-Optimal task, \textit{BankHeist}.
\begin{table}[h]                   
  \centering
  \footnotesize
  \caption{HNS Performance on visually bottlenecked tasks. }
  \label{tab:visual_bottleneck}
  \resizebox{\columnwidth}{!}{
    \begin{tabular}{
        l                                % Game
        rrrrrrrrrr                   % 13 numeric columns + HN
    }
\toprule
Game                &  TWM      &  DreamerV3    &  STORM    & IRIS                &  DIAMOND            &  JEDI(Ours)                          \\
\midrule
Breakout            &  {0.635}  &  {1.02}       &  {0.493}  & {2.85}              &  \underline{4.54}              &  \textbf{5.35}    \\
Assault             &  {0.886}  &  {0.931}      &  {1.11}   & \textbf{2.51}       &  \textbf{2.51}                 &  \underline{2.26}             \\
      \bottomrule
    \end{tabular}
    }
\end{table}

\textbf{Visually Bottlenecked Tasks}. Since it is difficult to quantitatively determine which tasks have high visual detail, we instead use the relative outperformance of pixel-based agents to identify tasks whose performance is limited by visual detail.  We define visually bottlenecked tasks as tasks where the HNS of both pixel-based agents (IRIS and DIAMOND) outperform the max over other baselines by more than 100\% (\cref{tab:visual_bottleneck}). JEDI achieves SOTA on \textit{Breakout} while being competitive in \textit{Assault}, defying the trend of underperformance in latent agents—answering our second research question. 
\begin{wrapfigure}{r}{0.5\columnwidth}
    \centering
    \includegraphics[width=0.5\columnwidth]{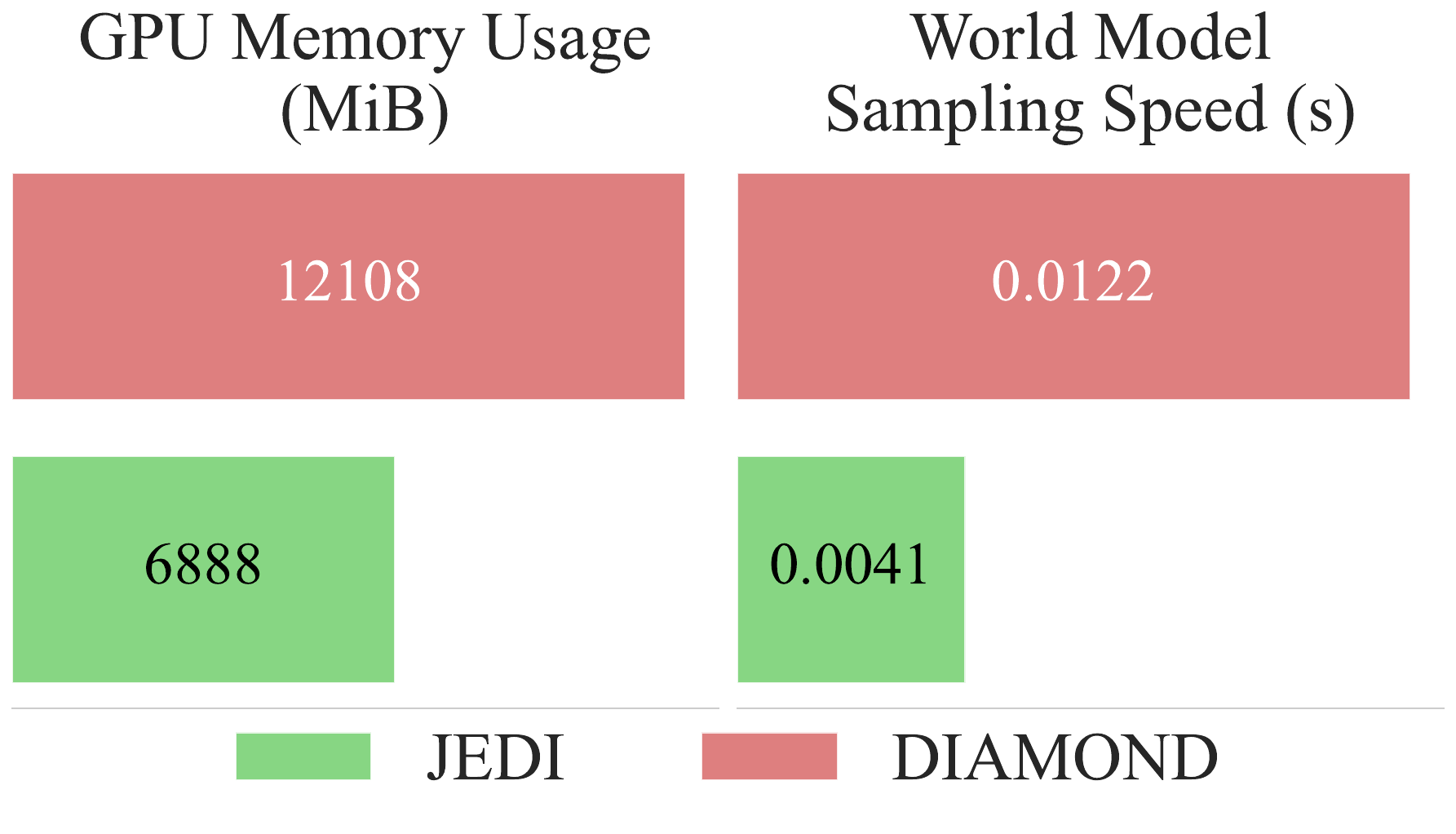}
    \caption{JEDI uses 57\% of DIAMOND's GPU Memory Usage while having 3$\times$ sampling speed. }
    \label{fig:compute_efficiency}
    % \vspace*{-4mm}
\end{wrapfigure}
This drives JEDI's overall performance on the Agent-Optimal subset as both tasks are Agent-Optimal. 

% Table \ref{tab:atari100k_aggregate_statistics} shows that JEDI achieves SOTA results on the Human-Optimal tasks, outperforming all baselines in mean, and interquartile mean (IQM), number of superhuman performances, and optimality gap. Remarkably, JEDI is the first lookahead-free method to achieve super-human performance in \textit{BankHeist} from the human-optimal set. On agent-optimal sets, JEDI retains runner-up performance across all metrics except optimality gap and number of superhuman performances. Results in \cref{fig:mbrl_vs_human_rawscore} further illustrates that JEDI is pushing the boundary between Agent-Optimal and Human-Optimal tasks rightwards. The results highlight its balanced performance, bridging the gap between human and agent performance.
% \subsection{Atari100k results}
\textbf{Atari100k Results.} On the full benchmark, JEDI achieves SOTA in number of super-human tasks and runner-up in mean, IQM and optimality gap (\cref{fig:jedi32_vs_baselines_full}). On Agent-Optimal tasks, JEDI achieves runner-up in number of super-human tasks, mean and IQM (\cref{fig:jedi32_vs_baselines_AO}). Thus, showing that JEDI achieves competitive results on conventional aggregates as well.

Remarkably, JEDI achieve these results with over 2$\times$ faster training (\cref{fig:compute_and_runtime_comparison}), over 3$\times$ faster inference, and 43\% less GPU memory usage compared to DIAMOND (\cref{fig:compute_efficiency}). This is due to the 12$\times$ latent compression achieved with JEDI.

\begin{figure}
    \centering
    \includegraphics[width=0.8\columnwidth]{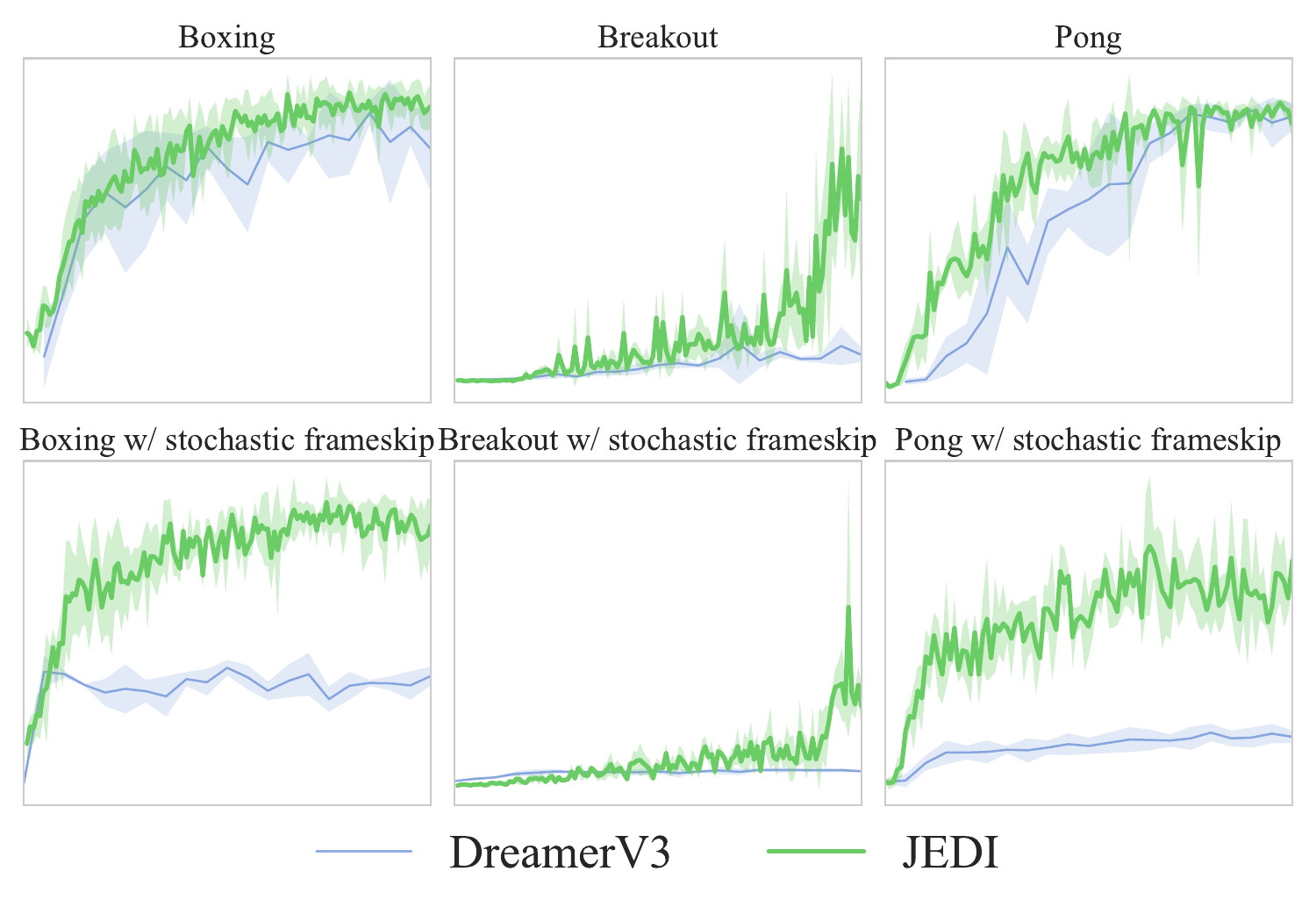}
    \caption{\small Bottom row is shows performance on same environment as top row, except with random frameskip. JEDI consistently outperform SOTA latent-based agent in stochastic settings.}
    \label{fig:stoch_experiments}
\end{figure}

% with frameskipping comparing JEDI and DreamerV3 on 3 Atari100k tasks
\textbf{Stochasticity.} Previous works demonstrated diffusion's prowess in modelling of stochastic multi-modal distributions \citep{chi2023DiffusionPolicy, janner2022diffuser}. As such, we tested JEDI in environments with random frameskipping. The results in Figure \ref{fig:stoch_experiments} supports the claim as JEDI outperforms SOTA latent non-diffusion agent substantially.
% , we turn our attention to Atari100k as a whole. Figure \ref{fig:jedi32_vs_baselines_HN} demonstrates that we achieve SOTA optimality gap, outperform DIAMOND on median score, and achieve second-best HNS mean across all baselines. Remarkably, we achieve these results with over 2 times faster training, over 3 times faster inference, and around 43\% less GPU memory usage compared to DIAMOND (Fig \ref{fig:compute_and_runtime_comparison}). This is solely due to the 12 times compression achieved with JEDI. 
\subsection{Ablation Study}
\begin{figure}[h]
    \centering
    \includegraphics[width=\columnwidth]{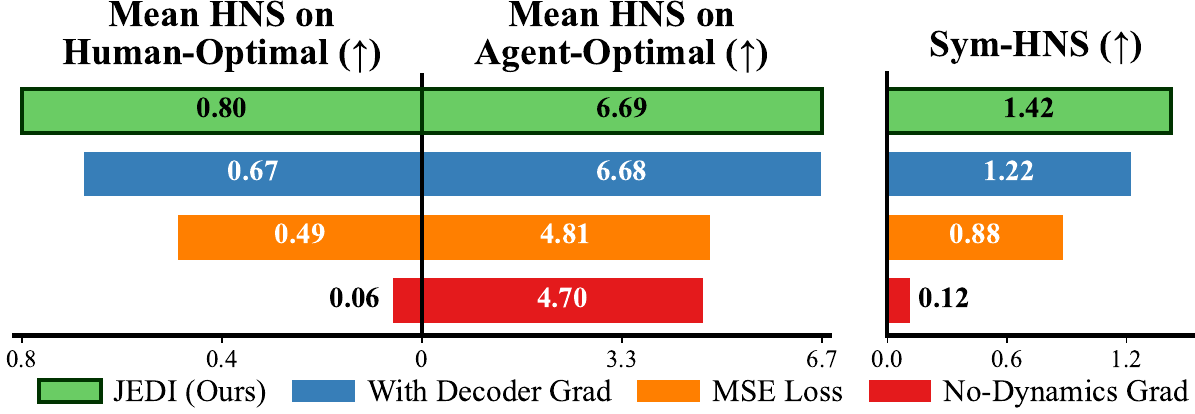}
    \caption{\small Ablation study on the effects of JEDI gradients on 3 tasks each from Agent/Human-Optimal tasks. Altering JEDI gradients hurts performance, especially for Human-Optimal tasks.}
    \label{fig:new_ablation_study}
\end{figure}
To validate the effectiveness of JEDI end-to-end latent diffusion denoising loss for latent representation learning, we ablated JEDI by (i) adding decoder gradients, (ii) replacing the diffusion denoising loss with a simple MSE loss, (iii) removing the diffusion denoising loss entirely (\cref{fig:new_ablation_study}). JEDI is effective, especially for Human-Optimal tasks. This can be seen from the decoder results, where JEDI ties performance on Agent-Optimal tasks and outperforms it in Human-Optimal tasks. Similarly, no-dynamics gradients perform decently on Agent-Optimal tasks but fail completely on Human-Optimal tasks compared to JEDI. We have further ablations to evaluate our technical design choices in \cref{fig:old_ablation_study}. 
\begin{figure*}[t]
    \centering
    \includegraphics[width=0.8\textwidth]{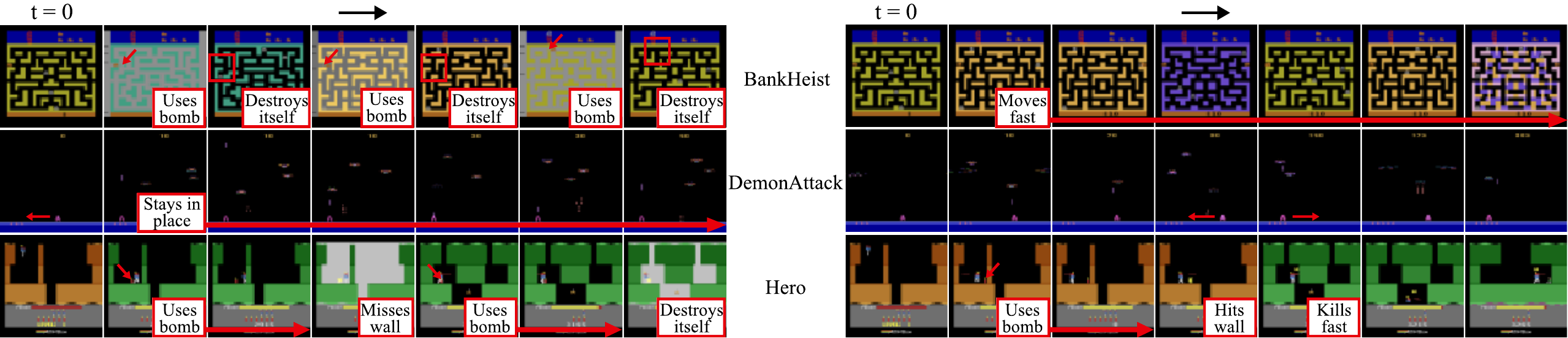}
    \caption{Example trajectories with DIAMOND (\textbf{LEFT}) and JEDI (\textbf{RIGHT}) on three tasks. JEDI is significantly more effective at chasing and shooting enemies in \textit{DemonAttack} and destroying obstacles in \textit{Hero} as compared to DIAMOND. In \textit{BankHeist}, both models shuffle between maps to rob banks that spawn at the exit for easy points. However, JEDI moves much faster and gets more rewards. In both \textit{BankHeist} and \textit{Hero}, JEDI's effectively minimizes self-destruction, whereas DIAMOND frequently self-destructs.}
    \label{fig:qualitative_analysis}
\end{figure*}
\setlength{\tabcolsep}{1pt}
% We observe that JEDI is better at temporal causal action reasoning for maximizing short-term rewards.
% We show the initial observation at $t=0$ and six salient frames from a single trajectory.
% \vspace{20pt}
\subsection{Qualitative Analysis}
\label{sec:qualitative_analysis}
We support our proposed analysis linking curse of dimensionality and task complexity with a qualitative study on \textit{BankHeist}, \textit{DemonAttack}, and \textit{Hero} (\cref{fig:qualitative_analysis}). They all have large action spaces except for  \textit{DemonAttack} and are all Shooter tasks. In \textit{BankHeist}, JEDI learns to repeatedly enter and exit maps, gaining easy points from banks that spawn near the exit. Crucially, JEDI does not use the \textit{FIRE} action that triggers self-destructing dynamite, achieving SOTA super-human performance. On the other hand, the DIAMOND agent learns this but repeatedly self-destructs with the \textit{FIRE} action, leading to the worst performance among the baselines. We observe similar behaviors in \textit{Hero} , where the DIAMOND agent fails to destroy obstacles and self-destructs more frequently. In \textit{DemonAttack}, DIAMOND learns an ineffective policy of staying at one corner and jittering while randomly using \textit{FIRE} at mostly nothing. JEDI chases enemies and neutralizes them effectively, achieving SOTA super-human performance. 

In all cases, DIAMOND struggles to reason about and deploy the \textit{FIRE} action appropriately. This can be attributed to the use of pixel inputs for its RL networks, which makes learning exponentially harder, especially in complex environments with high action space and complex Shooter dynamics. \textbf{By moving the diffusion dynamics model into latent space, JEDI is better able to use the \textit{FIRE} action in context, leading to drastically improved policies.}   

% In all three tasks, the effective policy for consistent short-term rewards involves destroying enemies or moving quickly (minimizing unnecessary actions that might be self-destructive), and JEDI does these well. Both outcomes reduce early termination, resulting in higher returns. 
% These examples show that the DIAMOND agent is unable to reason well about its actions into the near future. whereas the JEDI agent is able to do so. 
% We believe that this supports our hypothesis of connecting the pronounced performance asymmetry to the lack of a temporally structure latent space provided to DIAMOND.  
% \textbf{Key Insight.}\quad By causally predicting the next state, reward and termination, given the current state and action, JEDI encoder learns a well-structured latent space for effective causal action reasoning for future rewards. Every additional action adds exponential state-action-reward-termination sequence combinations and the pixel-based agent has to learn a latent state that is well-structured for these combinations from scratch with only the temporal difference learning objective \citep{sutton1988TDlearning}. 
% \vspace{-10pt}
% \vspace{-10pt}
\section{Conclusion}
We uncover the problem of Performance Asymmetry in MBRL, highlighting how conventional aggregates conceal task overfitting. We reveal this asymmetry by identifying a protected attribute—Agent/Human-Optimality—and propose a more balanced aggregate, Sym-HNS, that addresses the dominance of extreme values. An analysis of the two subsets exposes two distinct quantitative differences: action space and the prevalence of Shooter tasks. We trace the striking Performance Asymmetry in the SOTA agent to the curse of dimensionality associated with image inputs and the use of diffusion, which benefits visually demanding tasks. To this end, we propose JEDI, which leverages an end-to-end latent diffusion dynamics model, that reverses the Performance Asymmetry in the SOTA agent and demonstrates that diffusion in latent-based agents can aid performance on visually bottlenecked tasks (\cref{tab:claim_evidence_jedi}). 

Beyond addressing our motivating questions, our study opens several promising research directions. The roughly 10× gap between Agent/Human-Optimal tasks suggests a compelling avenue for deeper investigation: (i) understanding the fundamental characteristics of tasks in which humans excel versus those in which agents excel, and  (ii) the mechanisms within algorithms that drive success or failure on each subset. Finally, our findings motivate the exploration of Performance Asymmetry beyond Atari100k and MBRL more broadly.

\section{Related Works}
\label{related_works}
\textbf{MBRL}. Early world modelling can be attributed to Dyna~\citep{sutton1991DYNA}. \cite{ha2018World} proposed learning the world model in the latent space from pixels. \cite{hafner2019Planet} proposed learning a world model for planning . Dreamer~\citep{Hafner2020Dreamer} learns an actor-critic network using the representations learned by an end-to-end trained world model. DreamerV2~\citep{hafner2020Dreamerv2} introduced discrete latents.
DreamerV3~\citep{hafner2023Dreamerv3} offers improvements for domain generalization. In parallel, IRIS~\citep{micheli2022iris} developed a language model-inspired world model. Multiple works proposed the use of transformer: STORM~\citep{zhang2023storm}, and TWM~\citep{robine2023TWM}. \citet{chen2023latent,chen2022mirror} leveraged world models for modeling human belief states. \citet{hansen2022TDMPC, hansen2023tdmpc2} introduced the use of JEPA for learning world model dynamics. \citet{alonso2024diamond} introduced diffusion as world models for MBRL. \textbf{JEDI is the first end-to-end latent diffusion world model}.

\textbf{JEPA.} \citet{grill2020BYOL} originally proposed the use of a symmetric $L_2$ loss and hypothesized that it allows representation collapse to be an unstable solution. \citet{caron2021DINO} proposed cross-entropy loss and \citet{assran2023I-JEPA} used $L_2$ loss. \citet{bardes2024V-JEPA, assran2025vjepa2} proposed the use of $L_1$ loss, connecting this to the minimization of deviation from the Mean Absolute Median. \textbf{Our work shows that JEPA training paradigm is directly compatible with off-the-shelf diffusion loss}.

\textbf{Diffusion models} was first introduced by \citet{song2019NCSN, ho2020ddpm}. \citet{karras2022EDM} optimized the training framework. \citet{vahdat2021scorelatent} proposed latent diffusion trained jointly with reconstruction. Others leverage on pretraining  \citep{rombach2022StableDiffusion, parkerholder2024genie2}. It is effective for planning and generating data \citep{janner2022diffuser,lu2023synther, jackson2024pgd, ding2024diffusion}. \citet{chi2023DiffusionPolicy} uses diffusion models for joint policy and encoder learning. Diffusion models for representation learning involve searching through denoising outputs \citep{luo2023DiffHyperfeatures, yue2024ExpDiffTimeSteps}, leveraging distillation \citep{yang2023NUSDiffasRepLearner, zhang2022UnsupRepLearnPretrainedSEA}, pixel generation conditioned on compressed representation  \citep{pandey2022diffusevae, nielsen2023diffenc}, or using the network's hidden activations \citep{baranchuk2021LabelEfficientUNETFeat, xiang2023DiffAEAreUnified}. \textbf{Our work presents a novel latent diffusion model that jointly learns and predicts compressed representations using JEPA, effectively used for downstream RL task.}

\section{Impact Statement}
This paper presents work whose goal is to advance the field of machine learning. There are many potential societal consequences of our work, none of which we feel must be specifically highlighted here.
% \section{Discussion and limitations}
% \label{limitations}
% \vspace{-10pt}
% The scope of our work and the definitions proposed is limited to the Atari100k benchmark \citep{kaiser2019Atari100kSimPLE}. Even though JEDI achieves SOTA performance in the Human-Optimal tasks, the performance is still around one order of magnitude smaller than the performance on the Agent-Optimal tasks, indicating a wide gap in performance in these two tasks. Despite our method beating image-based diffusion methods in terms of computation speed, we are slower than existing latent MBRL baselines due to the nature of the multi-step denoising inference with every next state sampled. We are limited by the academic compute budget and hence only conducted the experiments with 3 seeds, and also limited to the small environment steps regime and small set of tasks in Atari100k, as opposed to the larger set of tasks in Atari Benchmark \citep{bellemare2013ALE}.

\bibliography{references}
\bibliographystyle{icml2026}

%%%%%%%%%%%%%%%%%%%%%%%%%%%%%%%%%%%%%%%%%%%%%%%%%%%%%%%%%%%%%%%%%%%%%%%%%%%%%%%
%%%%%%%%%%%%%%%%%%%%%%%%%%%%%%%%%%%%%%%%%%%%%%%%%%%%%%%%%%%%%%%%%%%%%%%%%%%%%%%
% APPENDIX
%%%%%%%%%%%%%%%%%%%%%%%%%%%%%%%%%%%%%%%%%%%%%%%%%%%%%%%%%%%%%%%%%%%%%%%%%%%%%%%
%%%%%%%%%%%%%%%%%%%%%%%%%%%%%%%%%%%%%%%%%%%%%%%%%%%%%%%%%%%%%%%%%%%%%%%%%%%%%%%
\newpage
\appendix
\onecolumn
\newpage
% \section{Technical Appendices and Supplementary Material}
% Technical appendices with additional results, figures, graphs and proofs may be submitted with the paper submission before the full submission deadline (see above), or as a separate PDF in the ZIP file below before the supplementary material deadline. There is no page limit for the technical appendices.
\section{Action Space and Shooter Tasks}
\begin{table}[h]  % 'r' means right side, width approx 55% text width
\renewcommand{\arraystretch}{0.8}
\centering
\caption{Agent-Optimal and Human-Optimal Atari100k tasks with Number of Actions and Shooter Task label.}
\label{tab:tasks_with_num_actions_shooter_task}
\resizebox{0.9\linewidth}{!}{%
\begin{tabular}{|l|c|c|l|c|c|}
\toprule
Agent-Optimal Tasks       & Num Actions   & Shooter Task?  & Human-Optimal Tasks          & Num Actions    & Shooter Task? \\
\midrule
\textit{Boxing}           & 18            & \xmark         & \textit{BankHeist}           & 18             & \cmark        \\
\textit{Krull}            & 18            & \xmark         & \textit{DemonAttack}         & 6              & \cmark        \\
\textit{CrazyClimber}     & 9             & \xmark         & \textit{Hero}                & 18             & \cmark        \\
\textit{Gopher}           & 8             & \xmark         & \textit{BattleZone}          & 18             & \cmark        \\
\textit{RoadRunner}       & 18            & \xmark         & \textit{Frostbite}           & 18             & \xmark        \\
\textit{Jamesbond}        & 18            & \cmark         & \textit{Qbert}               & 6              & \xmark        \\
\textit{Assault}          & 7             & \cmark         & \textit{MsPacman}            & 9              & \xmark        \\
\textit{Breakout}         & 4             & \xmark         & \textit{Asterix}             & 9              & \xmark        \\
\textit{KungFuMaster}     & 14            & \xmark         & \textit{ChopperCommand}      & 18             & \cmark        \\
\textit{Pong}             & 6             & \xmark         & \textit{Amidar}              & 10             & \xmark        \\
\textit{Kangaroo}         & 18            & \xmark         & \textit{Alien}               & 18             & \cmark        \\
\textit{UpNDown}          & 6             & \xmark         & \textit{PrivateEye}          & 18             & \xmark        \\
\textit{Freeway}          & 3             & \xmark         & \textit{Seaquest}            & 18             & \cmark        \\
\bottomrule
\end{tabular}
}
\renewcommand{\arraystretch}{1}
\end{table}

\section{Discussion on Other Aggregates}
\label{sec:discussion_other_aggregates}
\subsection{Further Discussion on Inter-Quartile Mean (IQM) and Optimality Gap}
Given the reasonable confidence intervals shown in \cref{fig:jedi32_vs_baselines_full}, and the drastic differences in score magnitudes across tasks in \cref{fig:mbrl_vs_human_rawscore}, it is straightforward to infer that the IQM across all runs will eliminate entire task performances as a whole rather than targeting random seed anomalies across all tasks. This is explicitly shown in \cref{tab:top_50_seed_runs} and \cref{tab:bottom_50_seed_runs} where we see almost all 5 runs of the same tasks consistently being in the top 25 percentile (e.g. \textit{Boxing}, \textit{CrazyClimber}, \textit{Krull}) and the bottom 25 percentile (e.g. \textit{PrivateEye}, \textit{Seaquest}, \textit{Freeway}) This encourages the forgoing of performance improvements in tasks that lies in these two ranges, especially the bottom 25 percentile.

\begin{table}[h]
    \tiny
    \centering
\begin{tabular}{llllllllllll}
\toprule
Seed (TWM) & TWM & Seed (IRIS) & IRIS & Seed (DreamerV3) & DreamerV3 & Seed (STORM) & STORM & Seed (DIAMOND) & DIAMOND & Seed (JEDI(Ours)) & JEDI(Ours) \\
\midrule
Krull\_4 & \textbf{7.261} & Boxing\_1 & \textbf{7.175} & Gopher\_1 & \textbf{8.141} & Boxing\_3 & \textbf{7.533} & Boxing\_1 & \textbf{8.275} & Boxing\_1 & \textbf{8.239} \\
Boxing\_2 & \textbf{7.219} & Boxing\_4 & \textbf{6.950} & Krull\_4 & \textbf{7.838} & Boxing\_2 & \textbf{7.533} & Boxing\_3 & \textbf{8.117} & Boxing\_5 & \textbf{8.232} \\
Boxing\_1 & \textbf{7.118} & Krull\_5 & \textbf{6.637} & Krull\_5 & \textbf{7.482} & Krull\_4 & \textbf{7.352} & Krull\_4 & \textbf{7.253} & Boxing\_3 & \textbf{7.944} \\
Boxing\_5 & \textbf{6.732} & Boxing\_5 & \textbf{6.283} & Boxing\_3 & \textbf{7.138} & Krull\_1 & \textbf{7.128} & Krull\_5 & \textbf{7.109} & Krull\_2 & \textbf{7.461} \\
Boxing\_3 & \textbf{6.281} & Krull\_2 & \textbf{5.402} & Boxing\_1 & \textbf{6.950} & Krull\_2 & \textbf{6.875} & Krull\_1 & \textbf{7.063} & Boxing\_2 & \textbf{7.233} \\
Krull\_1 & \textbf{5.270} & Boxing\_3 & \textbf{4.642} & Boxing\_4 & \textbf{6.471} & Gopher\_1 & \textbf{6.484} & Boxing\_2 & \textbf{6.992} & Breakout\_3 & \textbf{7.094} \\
Boxing\_4 & \textbf{4.905} & Assault\_5 & \textbf{4.450} & Boxing\_5 & \textbf{6.075} & Boxing\_5 & \textbf{6.167} & Breakout\_5 & \textbf{6.906} & Krull\_4 & \textbf{6.901} \\
UpNDown\_4 & \textbf{4.852} & Krull\_1 & \textbf{4.168} & Krull\_2 & \textbf{5.674} & Boxing\_4 & \textbf{5.975} & Boxing\_5 & \textbf{6.842} & Krull\_3 & \textbf{6.495} \\
Krull\_2 & \textbf{4.012} & Krull\_4 & \textbf{4.130} & CrazyClimber\_2 & \textbf{4.424} & Boxing\_1 & \textbf{5.950} & Gopher\_4 & \textbf{6.306} & Boxing\_4 & \textbf{6.466} \\
Krull\_3 & \textbf{3.463} & Boxing\_2 & \textbf{4.100} & Krull\_3 & \textbf{4.402} & Krull\_3 & \textbf{5.790} & Krull\_3 & \textbf{6.127} & Krull\_1 & \textbf{6.408} \\
CrazyClimber\_2 & \textbf{3.213} & Breakout\_1 & \textbf{3.573} & Krull\_1 & \textbf{4.222} & Krull\_5 & \textbf{4.774} & Boxing\_4 & \textbf{5.942} & Breakout\_5 & \textbf{5.351} \\
CrazyClimber\_1 & \textbf{2.803} & Krull\_3 & \textbf{3.169} & Boxing\_2 & \textbf{4.082} & Gopher\_4 & \textbf{4.665} & Krull\_2 & \textbf{5.292} & Breakout\_2 & \textbf{5.070} \\
CrazyClimber\_3 & \textbf{2.656} & Breakout\_5 & \textbf{3.000} & Kangaroo\_1 & \textbf{3.905} & Gopher\_2 & \textbf{4.423} & Breakout\_2 & \textbf{5.108} & Krull\_5 & \textbf{5.060} \\
CrazyClimber\_5 & \textbf{2.373} & Breakout\_2 & \textbf{2.913} & CrazyClimber\_5 & \textbf{3.382} & CrazyClimber\_5 & \textbf{2.889} & Breakout\_1 & \textbf{4.247} & Breakout\_4 & \textbf{4.960} \\
Krull\_5 & \textbf{2.248} & CrazyClimber\_4 & \textbf{2.646} & CrazyClimber\_4 & \textbf{3.304} & Jamesbond\_1 & \textbf{2.743} & CrazyClimber\_4 & \textbf{4.209} & Breakout\_1 & \textbf{4.251} \\
RoadRunner\_3 & \textbf{1.994} & Assault\_4 & \textbf{2.567} & RoadRunner\_1 & \textbf{3.186} & CrazyClimber\_2 & \textbf{2.725} & RoadRunner\_3 & \textbf{3.771} & Assault\_2 & \textbf{3.233} \\
RoadRunner\_5 & \textbf{1.855} & Assault\_1 & \textbf{2.423} & RoadRunner\_4 & \textbf{3.030} & RoadRunner\_1 & \textbf{2.715} & Breakout\_4 & \textbf{3.691} & CrazyClimber\_5 & \textbf{2.973} \\
Jamesbond\_4 & \textbf{1.412} & Breakout\_4 & \textbf{2.382} & Jamesbond\_5 & \textbf{2.770} & Kangaroo\_4 & \textbf{2.463} & CrazyClimber\_5 & \textbf{3.666} & CrazyClimber\_1 & \textbf{2.802} \\
Jamesbond\_3 & \textbf{1.382} & Breakout\_3 & \textbf{2.361} & CrazyClimber\_1 & \textbf{2.602} & RoadRunner\_5 & \textbf{2.400} & CrazyClimber\_2 & \textbf{3.532} & CrazyClimber\_3 & \textbf{2.500} \\
KungFuMaster\_3 & \textbf{1.340} & CrazyClimber\_2 & \textbf{2.210} & Jamesbond\_1 & \textbf{2.085} & Kangaroo\_3 & \textbf{2.343} & Gopher\_5 & \textbf{3.503} & RoadRunner\_4 & \textbf{2.478} \\
KungFuMaster\_5 & \textbf{1.332} & RoadRunner\_3 & \textbf{2.182} & Gopher\_5 & \textbf{1.737} & RoadRunner\_2 & \textbf{2.324} & CrazyClimber\_3 & \textbf{3.423} & RoadRunner\_2 & \textbf{2.478} \\
BankHeist\_4 & \textbf{1.273} & CrazyClimber\_1 & \textbf{2.129} & Breakout\_3 & \textbf{1.618} & RoadRunner\_4 & \textbf{2.216} & Kangaroo\_1 & \textbf{3.292} & Assault\_1 & \textbf{2.428} \\
Jamesbond\_5 & \textbf{1.264} & CrazyClimber\_3 & \textbf{2.074} & Assault\_5 & \textbf{1.533} & Gopher\_5 & \textbf{2.101} & RoadRunner\_2 & \textbf{3.218} & CrazyClimber\_4 & \textbf{2.317} \\
Jamesbond\_1 & \textbf{1.258} & Assault\_3 & \textbf{1.731} & BankHeist\_5 & \textbf{1.507} & CrazyClimber\_4 & \textbf{1.983} & Assault\_4 & \textbf{3.020} & Assault\_3 & \textbf{2.241} \\
Pong\_4 & \textbf{1.181} & Jamesbond\_1 & \textbf{1.675} & Gopher\_3 & \textbf{1.456} & CrazyClimber\_3 & \textbf{1.915} & CrazyClimber\_1 & \textbf{2.812} & RoadRunner\_3 & \textbf{2.048} \\
Pong\_5 & \textbf{1.181} & Jamesbond\_3 & \textbf{1.669} & RoadRunner\_3 & \textbf{1.390} & CrazyClimber\_1 & \textbf{1.666} & Breakout\_3 & \textbf{2.753} & Assault\_5 & \textbf{2.006} \\
BankHeist\_3 & \textbf{1.158} & Jamesbond\_5 & \textbf{1.654} & BankHeist\_4 & \textbf{1.307} & Jamesbond\_4 & \textbf{1.629} & Assault\_3 & \textbf{2.725} & Gopher\_1 & \textbf{1.942} \\
Pong\_1 & \textbf{1.146} & KungFuMaster\_5 & \textbf{1.624} & Jamesbond\_3 & \textbf{1.264} & Jamesbond\_5 & \textbf{1.611} & Assault\_1 & \textbf{2.595} & Gopher\_4 & \textbf{1.905} \\
CrazyClimber\_4 & \textbf{1.139} & Jamesbond\_4 & \textbf{1.589} & KungFuMaster\_3 & \textbf{1.170} & RoadRunner\_3 & \textbf{1.548} & RoadRunner\_4 & \textbf{2.580} & Gopher\_5 & \textbf{1.891} \\
Gopher\_3 & \textbf{1.137} & KungFuMaster\_3 & \textbf{1.519} & Pong\_1 & \textbf{1.153} & Jamesbond\_2 & \textbf{1.538} & Assault\_2 & \textbf{2.116} & Gopher\_2 & \textbf{1.810} \\
KungFuMaster\_4 & \textbf{1.132} & DemonAttack\_3 & \textbf{1.504} & Pong\_2 & \textbf{1.153} & Assault\_2 & \textbf{1.501} & Assault\_5 & \textbf{2.092} & Jamesbond\_5 & \textbf{1.686} \\
Pong\_3 & \textbf{1.128} & RoadRunner\_1 & \textbf{1.448} & Pong\_4 & \textbf{1.118} & KungFuMaster\_5 & \textbf{1.458} & RoadRunner\_5 & \textbf{2.059} & BankHeist\_5 & \textbf{1.662} \\
KungFuMaster\_1 & \textbf{1.074} & Assault\_2 & \textbf{1.358} & Pong\_5 & \textbf{1.110} & KungFuMaster\_2 & \textbf{1.312} & Kangaroo\_2 & \textbf{1.946} & Jamesbond\_2 & \textbf{1.600} \\
\midrule
Freeway\_4 & 1.052 & DemonAttack\_4 & 1.353 & KungFuMaster\_5 & 1.110 & BankHeist\_5 & 1.248 & Kangaroo\_3 & 1.897 & BankHeist\_1 & 1.595 \\
Freeway\_3 & 1.051 & Jamesbond\_2 & 1.333 & Jamesbond\_4 & 1.104 & Jamesbond\_3 & 1.245 & Gopher\_2 & 1.893 & BankHeist\_2 & 1.575 \\
Assault\_5 & 1.033 & Gopher\_3 & 1.324 & Breakout\_1 & 1.098 & Pong\_1 & 1.176 & Jamesbond\_3 & 1.751 & Jamesbond\_1 & 1.539 \\
Freeway\_2 & 1.011 & RoadRunner\_5 & 1.271 & CrazyClimber\_3 & 1.079 & Pong\_2 & 1.167 & RoadRunner\_1 & 1.560 & Jamesbond\_3 & 1.538 \\
Assault\_1 & 1.008 & Gopher\_1 & 1.267 & Pong\_3 & 1.068 & Pong\_4 & 1.164 & Jamesbond\_5 & 1.474 & Jamesbond\_4 & 1.518 \\
Assault\_3 & 1.002 & Pong\_1 & 1.133 & UpNDown\_5 & 1.058 & Assault\_1 & 1.154 & Jamesbond\_1 & 1.443 & Gopher\_3 & 1.495 \\
Freeway\_1 & 0.986 & Pong\_2 & 1.110 & RoadRunner\_2 & 1.036 & KungFuMaster\_1 & 1.143 & Jamesbond\_4 & 1.311 & DemonAttack\_4 & 1.375 \\
Pong\_2 & 0.957 & Freeway\_5 & 1.068 & KungFuMaster\_2 & 1.034 & Kangaroo\_2 & 1.096 & Jamesbond\_2 & 1.296 & Assault\_4 & 1.370 \\
Gopher\_4 & 0.922 & Freeway\_2 & 1.064 & Kangaroo\_2 & 1.022 & BankHeist\_4 & 1.068 & KungFuMaster\_2 & 1.180 & RoadRunner\_5 & 1.334 \\
Breakout\_5 & 0.836 & Pong\_3 & 1.057 & Jamesbond\_2 & 1.020 & Assault\_3 & 1.012 & Pong\_3 & 1.178 & DemonAttack\_5 & 1.240 \\
UpNDown\_5 & 0.825 & Freeway\_1 & 1.044 & Assault\_2 & 1.020 & UpNDown\_4 & 1.005 & Pong\_2 & 1.178 & Pong\_4 & 1.150 \\
RoadRunner\_1 & 0.822 & Freeway\_4 & 1.041 & KungFuMaster\_1 & 1.014 & Assault\_5 & 0.998 & Pong\_4 & 1.173 & Pong\_1 & 1.135 \\
Jamesbond\_2 & 0.772 & Freeway\_3 & 1.030 & KungFuMaster\_4 & 1.001 & KungFuMaster\_3 & 0.949 & Pong\_5 & 1.153 & Pong\_3 & 1.105 \\
Assault\_4 & 0.761 & Gopher\_2 & 1.030 & UpNDown\_2 & 0.933 & KungFuMaster\_4 & 0.905 & KungFuMaster\_3 & 1.143 & Pong\_5 & 1.104 \\
Breakout\_2 & 0.752 & DemonAttack\_2 & 1.022 & RoadRunner\_5 & 0.922 & Assault\_4 & 0.903 & Freeway\_1 & 1.142 & Freeway\_1 & 1.057 \\
Frostbite\_5 & 0.724 & RoadRunner\_4 & 0.971 & Assault\_1 & 0.865 & Gopher\_3 & 0.848 & Freeway\_2 & 1.139 & RoadRunner\_1 & 1.030 \\
Kangaroo\_5 & 0.666 & Pong\_5 & 0.938 & Frostbite\_4 & 0.784 & BankHeist\_1 & 0.831 & Freeway\_5 & 1.139 & Pong\_2 & 1.012 \\
Breakout\_3 & 0.655 & DemonAttack\_5 & 0.880 & Gopher\_2 & 0.744 & BankHeist\_3 & 0.797 & Pong\_1 & 1.133 & DemonAttack\_2 & 0.992 \\
Assault\_2 & 0.624 & Pong\_4 & 0.762 & Gopher\_4 & 0.677 & UpNDown\_2 & 0.786 & Freeway\_4 & 1.115 & KungFuMaster\_4 & 0.949 \\
RoadRunner\_2 & 0.608 & Kangaroo\_1 & 0.737 & Breakout\_2 & 0.670 & Breakout\_5 & 0.774 & Freeway\_3 & 1.095 & BankHeist\_3 & 0.921 \\
Breakout\_4 & 0.580 & KungFuMaster\_2 & 0.698 & Breakout\_5 & 0.648 & Kangaroo\_5 & 0.740 & Kangaroo\_4 & 1.075 & DemonAttack\_3 & 0.879 \\
UpNDown\_2 & 0.568 & CrazyClimber\_5 & 0.631 & Kangaroo\_4 & 0.550 & Pong\_5 & 0.646 & Gopher\_3 & 1.049 & DemonAttack\_1 & 0.722 \\
Gopher\_1 & 0.530 & KungFuMaster\_4 & 0.628 & Kangaroo\_3 & 0.544 & UpNDown\_3 & 0.598 & Asterix\_3 & 0.859 & BattleZone\_1 & 0.611 \\
RoadRunner\_4 & 0.528 & Gopher\_5 & 0.546 & Assault\_4 & 0.532 & Breakout\_1 & 0.583 & Kangaroo\_5 & 0.724 & KungFuMaster\_3 & 0.603 \\
KungFuMaster\_2 & 0.526 & UpNDown\_2 & 0.433 & Breakout\_4 & 0.514 & UpNDown\_5 & 0.555 & KungFuMaster\_4 & 0.636 & UpNDown\_5 & 0.532 \\
Kangaroo\_2 & 0.514 & Gopher\_4 & 0.424 & UpNDown\_1 & 0.506 & Frostbite\_5 & 0.507 & KungFuMaster\_1 & 0.626 & Kangaroo\_3 & 0.532 \\
Frostbite\_3 & 0.504 & BattleZone\_1 & 0.420 & UpNDown\_3 & 0.491 & Breakout\_2 & 0.497 & KungFuMaster\_5 & 0.520 & KungFuMaster\_1 & 0.498 \\
UpNDown\_1 & 0.379 & DemonAttack\_1 & 0.414 & BankHeist\_2 & 0.473 & MsPacman\_4 & 0.463 & UpNDown\_3 & 0.445 & BattleZone\_2 & 0.489 \\
Gopher\_2 & 0.373 & BattleZone\_2 & 0.351 & BankHeist\_3 & 0.451 & MsPacman\_5 & 0.463 & UpNDown\_5 & 0.444 & BankHeist\_4 & 0.418 \\
Kangaroo\_1 & 0.361 & KungFuMaster\_1 & 0.313 & BattleZone\_4 & 0.435 & Frostbite\_4 & 0.460 & Qbert\_4 & 0.367 & KungFuMaster\_2 & 0.417 \\
Hero\_4 & 0.360 & UpNDown\_1 & 0.292 & Hero\_3 & 0.418 & BattleZone\_2 & 0.435 & Qbert\_3 & 0.342 & UpNDown\_2 & 0.364 \\
Breakout\_1 & 0.354 & BattleZone\_5 & 0.289 & Hero\_2 & 0.412 & BattleZone\_5 & 0.426 & Asterix\_1 & 0.340 & Hero\_5 & 0.362 \\
\bottomrule
\end{tabular}
\caption{Top 50 percentile seed runs sorted in descending order for individual methods}
\label{tab:top_50_seed_runs}
\end{table}

\begin{table}[h]
    \tiny
    \centering
\begin{tabular}{ll|ll|ll|ll|ll|ll}
\toprule
Seed (TWM) & TWM & Seed (IRIS) & IRIS & Seed (DreamerV3) & DreamerV3 & Seed (STORM) & STORM & Seed (DIAMOND) & DIAMOND & Seed (JEDI(Ours)) & JEDI(Ours) \\
\midrule
Frostbite\_4 & 0.330 & UpNDown\_4 & 0.278 & Hero\_5 & 0.403 & Hero\_3 & 0.419 & Gopher\_1 & 0.336 & Asterix\_4 & 0.345 \\
Gopher\_5 & 0.326 & RoadRunner\_2 & 0.258 & DemonAttack\_3 & 0.402 & Hero\_4 & 0.417 & Qbert\_1 & 0.327 & UpNDown\_3 & 0.333 \\
Qbert\_4 & 0.315 & BattleZone\_3 & 0.257 & BankHeist\_1 & 0.381 & Hero\_5 & 0.417 & Asterix\_5 & 0.319 & BattleZone\_3 & 0.313 \\
Qbert\_5 & 0.308 & Kangaroo\_3 & 0.234 & Assault\_3 & 0.355 & UpNDown\_1 & 0.394 & MsPacman\_4 & 0.318 & BattleZone\_5 & 0.309 \\
UpNDown\_3 & 0.298 & Hero\_4 & 0.230 & UpNDown\_4 & 0.350 & Breakout\_3 & 0.392 & Asterix\_4 & 0.316 & KungFuMaster\_5 & 0.306 \\
Qbert\_2 & 0.283 & Kangaroo\_2 & 0.228 & BattleZone\_3 & 0.320 & Pong\_3 & 0.377 & Qbert\_5 & 0.315 & Kangaroo\_2 & 0.304 \\
Kangaroo\_4 & 0.267 & BattleZone\_4 & 0.221 & Qbert\_5 & 0.320 & Frostbite\_1 & 0.371 & UpNDown\_1 & 0.287 & Asterix\_1 & 0.295 \\
Qbert\_1 & 0.250 & UpNDown\_3 & 0.216 & Qbert\_1 & 0.314 & Qbert\_2 & 0.365 & Qbert\_2 & 0.280 & Qbert\_2 & 0.288 \\
BankHeist\_5 & 0.242 & ChopperCommand\_4 & 0.205 & BattleZone\_2 & 0.270 & BattleZone\_1 & 0.349 & Asterix\_2 & 0.270 & Qbert\_3 & 0.284 \\
BattleZone\_5 & 0.235 & Hero\_2 & 0.198 & Qbert\_4 & 0.245 & Qbert\_5 & 0.342 & MsPacman\_5 & 0.263 & Alien\_2 & 0.269 \\
MsPacman\_3 & 0.231 & Hero\_1 & 0.198 & Kangaroo\_5 & 0.245 & Qbert\_3 & 0.330 & MsPacman\_2 & 0.239 & Qbert\_5 & 0.267 \\
MsPacman\_5 & 0.218 & Hero\_3 & 0.192 & Hero\_4 & 0.244 & Kangaroo\_1 & 0.325 & UpNDown\_2 & 0.238 & Qbert\_4 & 0.261 \\
Hero\_2 & 0.217 & Hero\_5 & 0.191 & MsPacman\_5 & 0.243 & Qbert\_4 & 0.315 & BattleZone\_2 & 0.234 & Asterix\_2 & 0.250 \\
BankHeist\_2 & 0.215 & MsPacman\_5 & 0.171 & Qbert\_3 & 0.227 & BankHeist\_2 & 0.300 & MsPacman\_1 & 0.221 & BattleZone\_4 & 0.249 \\
Hero\_5 & 0.202 & MsPacman\_1 & 0.142 & BattleZone\_5 & 0.227 & Qbert\_1 & 0.288 & Hero\_4 & 0.221 & Asterix\_5 & 0.246 \\
ChopperCommand\_4 & 0.196 & UpNDown\_5 & 0.130 & Hero\_1 & 0.222 & MsPacman\_3 & 0.287 & Hero\_2 & 0.216 & Hero\_4 & 0.223 \\
MsPacman\_1 & 0.195 & ChopperCommand\_3 & 0.129 & Alien\_4 & 0.203 & MsPacman\_2 & 0.286 & MsPacman\_3 & 0.201 & Hero\_2 & 0.214 \\
Kangaroo\_3 & 0.184 & ChopperCommand\_5 & 0.117 & BattleZone\_1 & 0.191 & MsPacman\_1 & 0.281 & Hero\_5 & 0.189 & Hero\_3 & 0.207 \\
MsPacman\_2 & 0.176 & Amidar\_1 & 0.111 & MsPacman\_2 & 0.161 & BattleZone\_4 & 0.248 & DemonAttack\_5 & 0.171 & UpNDown\_1 & 0.205 \\
BankHeist\_1 & 0.175 & Asterix\_2 & 0.109 & DemonAttack\_4 & 0.147 & Hero\_1 & 0.225 & Amidar\_4 & 0.157 & UpNDown\_4 & 0.205 \\
DemonAttack\_5 & 0.165 & Kangaroo\_5 & 0.107 & Alien\_3 & 0.143 & Breakout\_4 & 0.219 & ChopperCommand\_3 & 0.149 & Qbert\_1 & 0.195 \\
ChopperCommand\_1 & 0.161 & Asterix\_3 & 0.105 & MsPacman\_4 & 0.141 & ChopperCommand\_1 & 0.216 & ChopperCommand\_4 & 0.138 & CrazyClimber\_2 & 0.187 \\
Hero\_3 & 0.154 & MsPacman\_2 & 0.095 & MsPacman\_1 & 0.134 & PrivateEye\_3 & 0.215 & Amidar\_2 & 0.133 & Kangaroo\_5 & 0.184 \\
ChopperCommand\_2 & 0.147 & Amidar\_3 & 0.093 & Asterix\_2 & 0.124 & PrivateEye\_1 & 0.208 & Amidar\_5 & 0.126 & MsPacman\_4 & 0.169 \\
MsPacman\_4 & 0.144 & Amidar\_4 & 0.092 & Alien\_2 & 0.118 & Hero\_2 & 0.203 & Amidar\_1 & 0.122 & Asterix\_3 & 0.142 \\
DemonAttack\_2 & 0.134 & BankHeist\_3 & 0.073 & MsPacman\_3 & 0.111 & Alien\_4 & 0.199 & BattleZone\_1 & 0.114 & Alien\_1 & 0.134 \\
ChopperCommand\_3 & 0.127 & ChopperCommand\_2 & 0.071 & Asterix\_1 & 0.102 & ChopperCommand\_3 & 0.199 & Amidar\_3 & 0.103 & ChopperCommand\_4 & 0.132 \\
Asterix\_1 & 0.125 & Amidar\_2 & 0.070 & DemonAttack\_1 & 0.100 & ChopperCommand\_2 & 0.163 & Alien\_3 & 0.099 & MsPacman\_2 & 0.115 \\
DemonAttack\_4 & 0.123 & MsPacman\_4 & 0.070 & Asterix\_3 & 0.098 & ChopperCommand\_5 & 0.152 & DemonAttack\_4 & 0.098 & MsPacman\_3 & 0.102 \\
Asterix\_4 & 0.116 & Asterix\_1 & 0.067 & Amidar\_5 & 0.094 & Alien\_5 & 0.148 & Alien\_4 & 0.087 & MsPacman\_1 & 0.098 \\
Asterix\_3 & 0.112 & Qbert\_4 & 0.065 & Frostbite\_3 & 0.092 & BattleZone\_3 & 0.148 & ChopperCommand\_1 & 0.086 & Alien\_3 & 0.097 \\
Hero\_1 & 0.112 & BankHeist\_5 & 0.061 & DemonAttack\_5 & 0.090 & Amidar\_4 & 0.145 & DemonAttack\_3 & 0.079 & Kangaroo\_1 & 0.097 \\
\midrule
Asterix\_5 & \textbf{0.102} & Asterix\_4 & \textbf{0.061} & Amidar\_2 & \textbf{0.082} & Asterix\_4 & \textbf{0.134} & UpNDown\_4 & \textbf{0.076} & MsPacman\_5 & \textbf{0.095} \\
Alien\_2 & \textbf{0.097} & Frostbite\_3 & \textbf{0.054} & Alien\_1 & \textbf{0.082} & Amidar\_2 & \textbf{0.131} & Hero\_3 & \textbf{0.073} & Amidar\_3 & \textbf{0.093} \\
Amidar\_4 & \textbf{0.095} & Qbert\_1 & \textbf{0.052} & Amidar\_4 & \textbf{0.076} & Amidar\_3 & \textbf{0.112} & Hero\_1 & \textbf{0.071} & ChopperCommand\_1 & \textbf{0.092} \\
Asterix\_2 & \textbf{0.092} & ChopperCommand\_1 & \textbf{0.052} & Amidar\_1 & \textbf{0.072} & Amidar\_1 & \textbf{0.105} & Alien\_1 & \textbf{0.069} & Alien\_5 & \textbf{0.092} \\
Alien\_1 & \textbf{0.089} & Frostbite\_1 & \textbf{0.049} & Amidar\_3 & \textbf{0.071} & Asterix\_5 & \textbf{0.101} & Alien\_2 & \textbf{0.066} & ChopperCommand\_3 & \textbf{0.079} \\
Amidar\_2 & \textbf{0.087} & Qbert\_3 & \textbf{0.047} & PrivateEye\_1 & \textbf{0.071} & Asterix\_1 & \textbf{0.096} & Frostbite\_2 & \textbf{0.054} & Amidar\_1 & \textbf{0.074} \\
BattleZone\_4 & \textbf{0.081} & BankHeist\_2 & \textbf{0.047} & Alien\_5 & \textbf{0.071} & ChopperCommand\_4 & \textbf{0.090} & Alien\_5 & \textbf{0.054} & ChopperCommand\_5 & \textbf{0.073} \\
DemonAttack\_1 & \textbf{0.076} & Asterix\_5 & \textbf{0.046} & Qbert\_2 & \textbf{0.070} & Asterix\_3 & \textbf{0.087} & DemonAttack\_1 & \textbf{0.051} & Hero\_1 & \textbf{0.071} \\
Amidar\_3 & \textbf{0.068} & BankHeist\_4 & \textbf{0.044} & Asterix\_5 & \textbf{0.063} & Amidar\_5 & \textbf{0.087} & Frostbite\_3 & \textbf{0.050} & Amidar\_5 & \textbf{0.070} \\
BattleZone\_2 & \textbf{0.063} & Frostbite\_2 & \textbf{0.043} & Frostbite\_5 & \textbf{0.060} & Frostbite\_2 & \textbf{0.079} & Frostbite\_5 & \textbf{0.048} & ChopperCommand\_2 & \textbf{0.066} \\
Alien\_5 & \textbf{0.051} & Frostbite\_4 & \textbf{0.043} & Asterix\_4 & \textbf{0.058} & PrivateEye\_2 & \textbf{0.076} & Frostbite\_4 & \textbf{0.047} & Frostbite\_5 & \textbf{0.049} \\
Amidar\_5 & \textbf{0.050} & MsPacman\_3 & \textbf{0.043} & Frostbite\_2 & \textbf{0.048} & Asterix\_2 & \textbf{0.074} & ChopperCommand\_2 & \textbf{0.046} & Frostbite\_4 & \textbf{0.049} \\
Frostbite\_2 & \textbf{0.048} & Alien\_5 & \textbf{0.041} & Frostbite\_1 & \textbf{0.046} & Alien\_2 & \textbf{0.072} & Frostbite\_1 & \textbf{0.044} & Frostbite\_3 & \textbf{0.048} \\
DemonAttack\_3 & \textbf{0.046} & BankHeist\_1 & \textbf{0.038} & DemonAttack\_2 & \textbf{0.036} & Alien\_3 & \textbf{0.070} & BankHeist\_5 & \textbf{0.021} & Frostbite\_2 & \textbf{0.043} \\
Frostbite\_1 & \textbf{0.045} & Frostbite\_5 & \textbf{0.038} & Seaquest\_5 & \textbf{0.017} & Alien\_1 & \textbf{0.058} & Seaquest\_2 & \textbf{0.017} & Amidar\_4 & \textbf{0.043} \\
Alien\_4 & \textbf{0.044} & Amidar\_5 & \textbf{0.035} & Seaquest\_2 & \textbf{0.016} & PrivateEye\_5 & \textbf{0.058} & Seaquest\_3 & \textbf{0.017} & Frostbite\_1 & \textbf{0.043} \\
Alien\_3 & \textbf{0.044} & Alien\_2 & \textbf{0.033} & Seaquest\_4 & \textbf{0.011} & Frostbite\_3 & \textbf{0.047} & Seaquest\_5 & \textbf{0.013} & Amidar\_2 & \textbf{0.038} \\
ChopperCommand\_5 & \textbf{0.043} & Qbert\_2 & \textbf{0.031} & Seaquest\_1 & \textbf{0.010} & DemonAttack\_5 & \textbf{0.019} & BankHeist\_1 & \textbf{0.007} & Seaquest\_4 & \textbf{0.037} \\
Amidar\_1 & \textbf{0.038} & Alien\_3 & \textbf{0.030} & Seaquest\_3 & \textbf{0.010} & Seaquest\_3 & \textbf{0.016} & BankHeist\_3 & \textbf{0.007} & Alien\_4 & \textbf{0.033} \\
Qbert\_3 & \textbf{0.035} & Qbert\_5 & \textbf{0.024} & PrivateEye\_2 & \textbf{0.002} & DemonAttack\_4 & \textbf{0.012} & ChopperCommand\_5 & \textbf{0.006} & Seaquest\_3 & \textbf{0.030} \\
Seaquest\_3 & \textbf{0.019} & Alien\_1 & \textbf{0.020} & PrivateEye\_5 & \textbf{0.001} & Seaquest\_4 & \textbf{0.011} & Seaquest\_1 & \textbf{0.006} & Kangaroo\_4 & \textbf{0.030} \\
Seaquest\_1 & \textbf{0.018} & Seaquest\_2 & \textbf{0.019} & PrivateEye\_4 & \textbf{0.001} & Seaquest\_5 & \textbf{0.011} & Seaquest\_4 & \textbf{0.005} & Seaquest\_1 & \textbf{0.028} \\
Seaquest\_5 & \textbf{0.017} & Seaquest\_1 & \textbf{0.016} & PrivateEye\_3 & \textbf{0.001} & Seaquest\_2 & \textbf{0.009} & BankHeist\_2 & \textbf{0.003} & Seaquest\_2 & \textbf{0.015} \\
Seaquest\_2 & \textbf{0.016} & Alien\_4 & \textbf{0.015} & Freeway\_5 & \textbf{0.000} & Seaquest\_1 & \textbf{0.008} & PrivateEye\_3 & \textbf{0.003} & Seaquest\_5 & \textbf{0.014} \\
Seaquest\_4 & \textbf{0.014} & Seaquest\_3 & \textbf{0.015} & Freeway\_3 & \textbf{0.000} & DemonAttack\_2 & \textbf{0.005} & BattleZone\_4 & \textbf{0.001} & PrivateEye\_3 & \textbf{0.001} \\
BattleZone\_3 & \textbf{0.009} & Seaquest\_5 & \textbf{0.013} & Freeway\_4 & \textbf{0.000} & DemonAttack\_3 & \textbf{0.003} & PrivateEye\_1 & \textbf{0.001} & PrivateEye\_4 & \textbf{0.001} \\
BattleZone\_1 & \textbf{0.001} & Kangaroo\_4 & \textbf{0.013} & Freeway\_2 & \textbf{0.000} & PrivateEye\_4 & \textbf{0.001} & PrivateEye\_2 & \textbf{0.001} & PrivateEye\_5 & \textbf{0.001} \\
PrivateEye\_1 & \textbf{0.001} & Seaquest\_4 & \textbf{0.008} & Freeway\_1 & \textbf{0.000} & Freeway\_1 & \textbf{0.000} & PrivateEye\_5 & \textbf{0.001} & PrivateEye\_1 & \textbf{0.001} \\
PrivateEye\_2 & \textbf{0.001} & PrivateEye\_5 & \textbf{0.001} & ChopperCommand\_4 & \textbf{-0.040} & Freeway\_2 & \textbf{0.000} & PrivateEye\_4 & \textbf{0.001} & PrivateEye\_2 & \textbf{0.001} \\
PrivateEye\_3 & \textbf{0.001} & PrivateEye\_4 & \textbf{0.001} & ChopperCommand\_1 & \textbf{-0.050} & Freeway\_3 & \textbf{0.000} & BankHeist\_4 & \textbf{-0.001} & Freeway\_2 & \textbf{0.000} \\
PrivateEye\_4 & \textbf{0.001} & PrivateEye\_3 & \textbf{0.001} & ChopperCommand\_3 & \textbf{-0.055} & Freeway\_5 & \textbf{0.000} & BattleZone\_3 & \textbf{-0.004} & Freeway\_3 & \textbf{0.000} \\
PrivateEye\_5 & \textbf{0.000} & PrivateEye\_2 & \textbf{0.001} & ChopperCommand\_2 & \textbf{-0.078} & Freeway\_4 & \textbf{0.000} & BattleZone\_5 & \textbf{-0.009} & Freeway\_5 & \textbf{0.000} \\
Freeway\_5 & \textbf{0.000} & PrivateEye\_1 & \textbf{0.001} & ChopperCommand\_5 & \textbf{-0.081} & DemonAttack\_1 & \textbf{-0.004} & DemonAttack\_2 & \textbf{-0.025} & Freeway\_4 & \textbf{0.000} \\
\bottomrule
\end{tabular}
\caption{Bottom 50 percentile seed runs sorted in descending order for individual methods}
\label{tab:bottom_50_seed_runs}
\end{table}

Optimality Gap averages the mean gap of tasks to score HNS of 1. Unfortunately, it does not take into account the success of agents beyond HNS of 1.0.
\[
\text{Optimality Gap}
= \frac{1}{N} \sum_{n=1}^{N} \max\!\left(1 - \min\!\left(1,\,\max\!\left(0,\,HNS_n\right)\right),\,0\right).
\]

Both aggregates are not aligned with our intent of celebrating performance and all forms of progress across all tasks. Nonetheless, we still report these aggregates in our results.

\subsection{Median}
The use of Median in Atari was already criticized in \citet{agarwal2021precipice} as zero scores on nearly half the tasks do not affect it. This is antithetical to our work on alleviating Performance Asymmetry.

\subsection{Variance/Standard Deviation}
One natural aggregate to consider is using the variance/standard deviation of data to measure Performance Asymmetry. However, variance itself is derived using the arithmetic mean. This implies that it is subject to the same issues of extreme and dominant values. For instance, a few tasks having extremely high variance will negate the effects of many tasks being close to the mean. Indeed, from \cref{tab:mean_std_dev}, all of the standard deviations are relatively high and proportionate to their means. As such we are unable to decisively observe Performance Asymmetry seen in \cref{fig:T_plot_all_results} and \cref{fig:mbrl_vs_human_rawscore}. This is because \textbf{all of the MBRL agents perform exceptionally well on the same set of tasks with low human scores}, resulting in high standard deviation anomalies that mask meaningful gains in other tasks that reduce standard deviation. 
\begin{table}[H]
\centering
\caption{Mean and standard deviation of MBRL Agents.}
\begin{tabular}{lcc}
\toprule
Method & Mean & Std.\ Dev. \\
\midrule
TWM & 0.956 & 1.55 \\
IRIS & 1.05 & 1.53 \\
DreamerV3 & 1.13 & 1.81 \\
STORM & 1.22 & 1.83 \\
DIAMOND & 1.46 & 2.08 \\
JEDI (Ours) & 1.36 & 2.04 \\
\bottomrule
\end{tabular}
\label{tab:mean_std_dev}
\end{table}

Moreover, optimizing methods for low standard deviation of task scores has the adverse effects of punishing exceptional performances. This is again antithetical to our work as we want to celebrate the both the success and failure modes of MBRL as they contain rich avenues for future holistic advancements. 

\subsection{Harmonic Mean Over All Tasks in Atari100k}
This is unfeasible due to numerical reasons. The range of HNS lies within $[-0.08, 8.3]$. The prevalence score of $0$, for tasks like \textit{Freeway}, collapses the harmonic mean to 0, and negative scores renders it to be potentially unstable (dividing by 0) and meaningless. Hence, we overcome this numerical issue by first taking the arithmetic mean of HNS over Agent-Optimal and Human-Optimal subsets, enabling $>0$ mean HNS almost certainly in the current era of MBRL. 

\section{Supplemetary Background}
\subsection{Model-Based Reinforcement Learning}
MBRL attempts to solve Partially Observable Markov Decision Processes (POMDP) environments \citep{kaelbling1998POMDP} with RL where the agent does not have direct access to states, $z_t$, and only have access to observations containing partial information $o_t$. The objective is to maximize the expected cumulative sum of discounted rewards. This is typically done through learning a generative world model which consists of an encoder, dynamics model, reward and termination model with SSL. The RL agent is then learned within compressed latent trajectories of the world model, and the trained agent is deployed in the real environment to collect data for training the world model, and the cycle repeats.
\subsection{JEPA}
JEPA learns by training a predictor to map target embeddings of equivariant views of data modality in the same latent space. To avoid representation collapse---a problem that occurs when a representation learner tends to represent data with similar embeddings, modern day JEPA generally relies on data augmentation and self teacher-student training~\citep{caron2021DINO}, which is a result of a number of works.

\subsection{Score-based diffusion models and flow matching}
Modern diffusion models can be generalized to a continuous score-based generative SDE of a random variable $\{\mathbf{z}^\tau\}_{\tau \in [0,\mathcal{T}]}$ where $\tau \in [0, \mathcal{T}]$ is the diffusion time variable. We are interested in modeling the stochastic process between data and noise, and thus the marginal distributions of $\mathbf{z}$, $\{p^\tau \}_{\tau \in [0,\mathcal{T}]}$, begins with the data distribution $p^0$ and ends with a Gaussian, $p^\mathcal{T}$. The forward process from data to noise is modeled as a diffusion process, and it turns out that the reverse process, from noise to data is also a  diffusion process, given by the following SDEs :
a formalization of adding cumulative infinitesimal Gaussian noise over time
learning and optimal score model appears to involves discretising and solving of the forward and backward SDE
Modern diffusion models can be generalized to a continuous score-based generative SDE which flows forward and backward from gaussian noise to data \citep{ANDERSON1982RevDiffusion, song2020ScoreBasedSDEs}:
\begin{equation}
    d\mathbf{z} = \mathbf{f}(\mathbf{z}, \tau)d\tau + g(\tau)d\mathbf{w}
\end{equation}
\begin{equation}
    d\mathbf{z} = [\mathbf{f}(\mathbf{z},\tau) - g(\tau)^2 \nabla_\mathbf{z}\log p^\tau(\mathbf{z})]d\tau + g(\tau)d\Bar{\mathbf{w}}
\end{equation}
where $\mathbf{w}$ is the Wiener process, and $g(\tau): \mathbb{R} \rightarrow \mathbb{R}$ is the amplitude of noise at time $\tau$.
$\mathbf{f}(.,\tau): \mathbb{R}^d \rightarrow \mathbb{R}^d$ is the drift coefficient (typically 0). $\nabla_\mathbf{z}\log p^\tau(\mathbf{z})$ is the score function that determines the direction from noise to data given $\tau$, which can be estimated with a neural network score model $\mathbf{F}_\theta(\mathbf{z}, \tau)$. 
Modern diffusion models can be generalized to a continuous score-based generative SDE which flows forward and backward from gaussian noise to data\citep{ANDERSON1982RevDiffusion, song2020ScoreBasedSDEs}. At first glance, learning a neural network $\textbf{F}_\theta$ for this resembles earlier works that require solving an ODE during training\citep{chen2018NeuralODE, rezende2015NormFlow}. However, the main novelty of diffusion models lies in the fact that simulation-free efficient and unbiased training can be achieved because a specific sample-conditional path was pre-determined. Surprisingly, the intractable objective of regressing $\mathbf{F}_\theta(\mathbf{z}, \tau)$ towards the marginal probability $\mathbf{p(z)}$ at diffusion time step $\tau$: 
\begin{equation}
    \mathcal{L}(\theta) = \mathbb{E}_{\tau, z^0} [||\mathbf{F_\theta(\mathbf{z^\tau},\tau) - \nabla_{\mathbf{z}^\tau}\log p^{\tau}(\mathbf{z}})||^2]
\end{equation}
has the same gradient as regressing towards the tractable sample-conditional vector field derived from our choice of probability path in expectation \citep{albergo2023StochInterpol, lipman2022FlowMatching} :
\begin{equation}
    \mathcal{L}(\theta) = \mathbb{E}_{\tau, z^0} [||\mathbf{F_\theta(\mathbf{z^\tau},\tau) - \nabla_{\mathbf{z}^\tau}\log p^{0\tau}(\mathbf{z}^\tau|\mathbf{z^0}})||^2]
    \label{eqn:flow_matching}
\end{equation}
where the expectation is over $\tau$, and the noised sample $\mathbf{z \sim p^{0\tau}(z^\tau | z^0)}$, depends on the pre-determined probability path conditional on clean samples $z^0 \sim p^0$. This can be derived from linear interpolation or the standard diffusion path at time step $\tau$. With this, the seond term for Eqn \ref{eqn:flow_matching} becomes trivial.

\section{Changes Compared to Baseline and Hyperparameters}
\label{appendix:hyperparameters}
We used the DIAMOND library \citep{alonso2024diamond} (MIT License) for our implementation. The main difference between JEDI World Model and DIAMOND is the transplant of the DIAMOND's RL encoder onto the JEDI World Model and the techniques implemented to facilitate end-to-end latent diffusion mentioned in \cref{sec:method_jedi}. The only network architectural changes are practical: (1) changing the original RL encoder to downsample to $z_t \in [-3,3]^{16\times8\times8}$, (2) modifying the diffusion UNET \citep{ronneberger2015UNET} into a single layer without downsampling, (3) and removing the downsampling in the Reward/Termination Model's encoder. The only hyperparameter changes were increasing the diffusion model's warm-up learning steps to 1000 and sigma of input data to be 1. 

\begin{table}[h]
\centering
\caption{Model architecture details for JEDI. Green font refers to changes as compared to pixel-based diffusion world model baseline \citep{alonso2024diamond}. Downsampling layers refers to the Maxpool operation that downsamples the input height and width by a factor of 2.}
\begin{tabular}{l@{\hspace{2em}}c}
\toprule
\textbf{Hyperparameter} & \textbf{Value} \\
\midrule
\multicolumn{2}{l}{\textbf{Latent Diffusion Dynamics Model ($\mathbf{D}_\theta$)}} \\
State conditioning mechanism & Frame stacking \\
Action conditioning mechanism & Adaptive Group Normalization \\
Diffusion time conditioning mechanism & Adaptive Group Normalization \\
Residual blocks layers & \textcolor{darkgreen}{[1]} \\
Residual blocks channels & \textcolor{darkgreen}{[160]} \\
UNET Downsampling &  \textcolor{darkgreen}{NIL} \\
Residual blocks conditioning dimension & 256 \\
Sigma data (for preconditioning) & \textcolor{darkgreen}{1} \\
\addlinespace
\textcolor{darkgreen}{\textbf{World Model Encoder ($\mathbf{E}_\phi$)}} \\
\textcolor{darkgreen}{{Encoder blocks layers}} & \textcolor{darkgreen}{[1, 1, 1, 1]} \\
\textcolor{darkgreen}{{Encoder blocks channels}} & \textcolor{darkgreen}{[32, 32, 32, 16]} \\
\textcolor{darkgreen}{Encoder Downsampling layers} &  \textcolor{darkgreen}{[1,1,1,0]} \\
\textcolor{darkgreen}{Encoder tanh clamp factor} & \textcolor{darkgreen}{3} \\
\addlinespace

\multicolumn{2}{l}{\textbf{Reward/Termination Model ($\mathbf{R}_\psi$)}} \\
Action conditioning mechanisms & Adaptive Group Normalization \\
Residual blocks layers & {[2, 2, 2, 2]} \\
Residual blocks channels & {[32, 32, 32, 32]} \\
Encoder Downsampling layers &  \textcolor{darkgreen}{NIL} \\
Residual blocks conditioning dimension & 128 \\
LSTM dimension & 512 \\
\addlinespace

\multicolumn{2}{l}{\textbf{Actor-Critic Model ($\pi_\omega$ and $V_\omega$)}} \\
Encoder & \textcolor{darkgreen}{NIL} \\
LSTM dimension & 512 \\
\bottomrule
\end{tabular}
\end{table}

\begin{table}[h]
\centering
\caption{Training hyperparameters for JEDI. Green font refers to changes as compared to pixel-based diffusion world model baseline \citep{alonso2024diamond}.}
\begin{tabular}{l@{\hspace{2em}}c}
\toprule
\textbf{Hyperparameter} & \textbf{Value} \\
\midrule

\multicolumn{2}{l}{\textbf{Training loop}} \\
Number of epochs & 1000 \\
Training steps per epoch & 400 \\
Batch size & 32 \\
Environment steps per epoch & 100 \\
Epsilon (greedy) for collection & 0.01 \\
\addlinespace

\multicolumn{2}{l}{\textbf{RL hyperparameters}} \\
Imagination horizon ($H$) & 15 \\
Discount factor ($\gamma$) & 0.985 \\
Entropy weight ($\eta$) & 0.001 \\
$\lambda$-returns coefficient ($\lambda$) & 0.95 \\
\addlinespace

\multicolumn{2}{l}{\textbf{Sequence construction during training}} \\
For $D_\theta$, number of conditioning observations and actions ($L$) & 4 \\
For $R_\psi$, burn-in length ($B_R$), set to $L$ in practice & 4 \\
For $R_\psi$, training sequence length ($B_R + H$) & 19 \\
For $\pi_\phi$ and $V_\phi$, burn-in length ($B_{\pi,V}$), set to $L$ in practice & 4 \\
\addlinespace

\multicolumn{2}{l}{\textbf{Optimization}} \\
Optimizer & AdamW \\
Learning rate & 1e-4 \\
Epsilon & 1e-8 \\
Weight decay ($D_\theta$) & 1e-2 \\
Weight decay ($R_\psi$) & 1e-2 \\
Weight decay ($\pi_\phi$ and $V_\phi$) & 0 \\
Learning rate Warm-up steps ($\mathbf{D}_\theta$) & \textcolor{darkgreen}{1e3} \\
Learning rate Warm-up steps($\mathbf{R}_\psi$) & 1e2 \\
Learning rate Warm-up steps ($\pi_\omega$ and $V_\omega$) & 1e2 \\
\textcolor{darkgreen}{Learning rate scale factor for $\mathbf{E_\phi}$} & \textcolor{darkgreen}{0.3} \\
\addlinespace

\multicolumn{2}{l}{\textbf{Diffusion Sampling}} \\
Method & Euler \\
Number of steps & 3 \\
S churn & \textcolor{darkgreen}{1 (only for stochastic experiments)}\\
\addlinespace

\multicolumn{2}{l}{\textbf{Environment}} \\
Image observation dimensions & $64 \times 64 \times 3$ \\
Action space & Discrete (up to 18 actions) \\
Frameskip & 4 \\
Frameskip for Stochastic Experiments & [2, 6] \\
Max noop & 30 \\
Termination on life loss & True \\
Reward clipping & \{-1, 0, 1\} \\
\bottomrule
\end{tabular}
\end{table}

\subsection{JEDI Algorithm}
Green font refers to changes as compared to pixel-based diffusion world model baseline \citep{alonso2024diamond}
\begin{algorithm}
\footnotesize
\caption{JEDI Training}
\label{alg:jedi}
\begin{algorithmic}[1]
\Procedure{training\_loop}{}
    \For{epochs}
        \State collect\_experience(steps\_collect)
        \For{steps\_diffusion\_model}
            \State \textcolor{darkgreen}{update\_latent\_diffusion\_model()}
        \EndFor
        \For{steps\_reward\_end\_model}
            \State update\_reward\_end\_model()
        \EndFor
        \For{steps\_actor\_critic}
            \State update\_actor\_critic()
        \EndFor
    \EndFor
\EndProcedure
\Statex
\Procedure{collect\_experience}{$n$}
    \State $x_0 \leftarrow \text{env.reset}()$
    \For{$t=0$ \textbf{to} $n-1$}
        \State Sample $a_t \sim \textcolor{darkgreen}{\pi_\omega(a_t | z_t^0)q_\phi(z_t^0|x_t)}$ \Comment{derive $z_t^0$ using JEDI encoder $\mathbf{E}_\phi$ before sampling action}
        \State $x_{t+1}, r_t, d_t \leftarrow \text{env.step}(a_t)$
        \State $\mathcal{D} \leftarrow \mathcal{D} \cup \{x_t, a_t, r_t, d_t\}$
        \If{$d_t = \text{true}$}
            \State $x_{t+1} \leftarrow \text{env.reset}()$
        \EndIf
    \EndFor
\EndProcedure
\Statex
\Procedure{\textcolor{darkgreen}{update\_latent\_diffusion\_model}}{}
    \State Sample sequence $(x_{t-L+1}, a_{t-L+1}, \dots, x_t, a_t, x_{t+1}) \sim \mathcal{D}$
    \State Sample $\log(\sigma) \sim \mathcal{N}(P_{\text{mean}}, P_{\text{std}}^2)$ \Comment{log-normal sigma distribution from EDM}
    \State Define $\tau := \sigma$ \Comment{default identity schedule from EDM}
    \State \textcolor{darkgreen}{compute $[z_{t-L+1}^0, \dots, z_t^0, z_{t+1}^0] = C(\mathbf{E}_\phi([x_{t-L+1}, \dots, x_t, x_{t+1}])$)} \Comment {derive the clamped latents}
    \State \textcolor{darkgreen}{derive $z_{t+1}^0 = \textnormal{sg(}z_{t+1}^0\textnormal{)}$ } \Comment {detach gradients from the target latent}
    \State Sample $\textcolor{darkgreen}{{z}_{t+1}^\tau} \sim \mathcal{N}(\textcolor{pastelgreen}{z_{t+1}^0}, \sigma^2 \mathbf{I})$ \Comment{add independent Gaussian noise}
    \State Compute \textcolor{darkgreen}{$\hat{z}_{t+1}^{0} = \mathbf{D}_\theta({z}_{t+1}^\tau, \tau, z_{k-L+1}^0, a_{t-L+1}, \dots, z_t^0, a_t)$}
    \State Compute reconstruction loss $\mathcal{L}(\theta) = \| \textcolor{darkgreen}{\hat{z}_{t+1}^{0} - z_{t+1}^0} \|^2$
    \State \textcolor{darkgreen}{$rs \;\sim\; \mathrm{Uniform}\{\,\text{True},\text{False}\,\}$} 
    \If{\textcolor{darkgreen}{$rs$ = True}} \Comment{random switch between $\hat{z}_{t+1}^0$ or ${z}_{t+1}^0$ as $\mathbf{D}_\theta$ input for loss at $t+1$}
        \State \textcolor{darkgreen}{Cache $\hat{z}_{t+1}^0$ as input to $\mathbf{D}_\theta$ for {\scriptsize UPDATE\_LATENT\_DIFFUSION\_MODEL} at $t+1$}
    \EndIf
    \State Update $\mathbf{D}_\theta$
\EndProcedure
\Statex
\Procedure{update\_reward\_end\_model}{}
    \State Sample indexes $\mathcal{I} = \{t, \dots, t+L+H-1\}$ \Comment{burn-in + imagination horizon}
    \State Sample sequence $(x_i, a_i, r_i, d_i)_{i \in \mathcal{I}} \sim \mathcal{D}$
    \State \textcolor{darkgreen}{Compute $(z_{i}^0)_{i\in \mathcal{I}} = C(\mathbf{E}_\phi((x_{i})_{i\in \mathcal{I}})$)} \Comment {derive the clamped latents}
    \State Initialize $h=c=0$ \Comment{LSTM hidden and cell states}
    \For{$i \in \mathcal{I}$ \textbf{do}}
        \State Compute $\hat{r}_i, \hat{d}_i, h, c = R_\psi(\textcolor{darkgreen}{z_i^0}, a_i, h, c)$
    \EndFor
    \State Compute $\mathcal{L}(\psi) = \sum_{i \in \mathcal{I}} \text{CE}(\hat{r}_i, \text{sign}(r_i)) + \text{CE}(\hat{d}_i, d_i)$ \Comment{CE: cross-entropy loss}
    \State Update $R_\psi$
\EndProcedure
\Statex
\Procedure{update\_actor\_critic}{}
    \State Sample initial buffer $(x_{t-L+1}, a_{t-L+1}, \dots, x_t) \sim \mathcal{D}$
    \State \textcolor{darkgreen}{Compute $[z_{t-L+1}^0, \dots, z_t^0] = C(\mathbf{E}_\phi([x_{t-L+1}, \dots, x_t])$)} \Comment {derive the clamped latents}
    \State Burn-in buffer with $R_\psi, \pi_\omega$ and $V_\omega$ to initialize LSTM states
    \For{$i=t$ \textbf{to} $t+H-1$}
        \State Sample $a_i \sim \pi_\omega(a_i | \textcolor{darkgreen}{z_i^0})$
        \State Sample reward $r_i$ and termination $d_i$ with $R_\psi$
        \State Sample next observation \textcolor{darkgreen}{$z_{i+1}^0$} by simulating reverse diffusion process with $\mathbf{D}_\theta$
    \EndFor
    \State Compute $V_\omega(\textcolor{darkgreen}{z_t^0})$ for $i = t, \dots, t+H$
    \State Compute RL losses $\mathcal{L}_V(\omega)$ and $\mathcal{L}_\pi(\omega)$
    \State Update $\pi_\omega$ and $V_\omega$
\EndProcedure
\end{algorithmic}
\end{algorithm}

\clearpage
\newpage
\section{Supplementary Results}
\begin{figure}[H]
  \centering
   \includegraphics[width=\columnwidth]{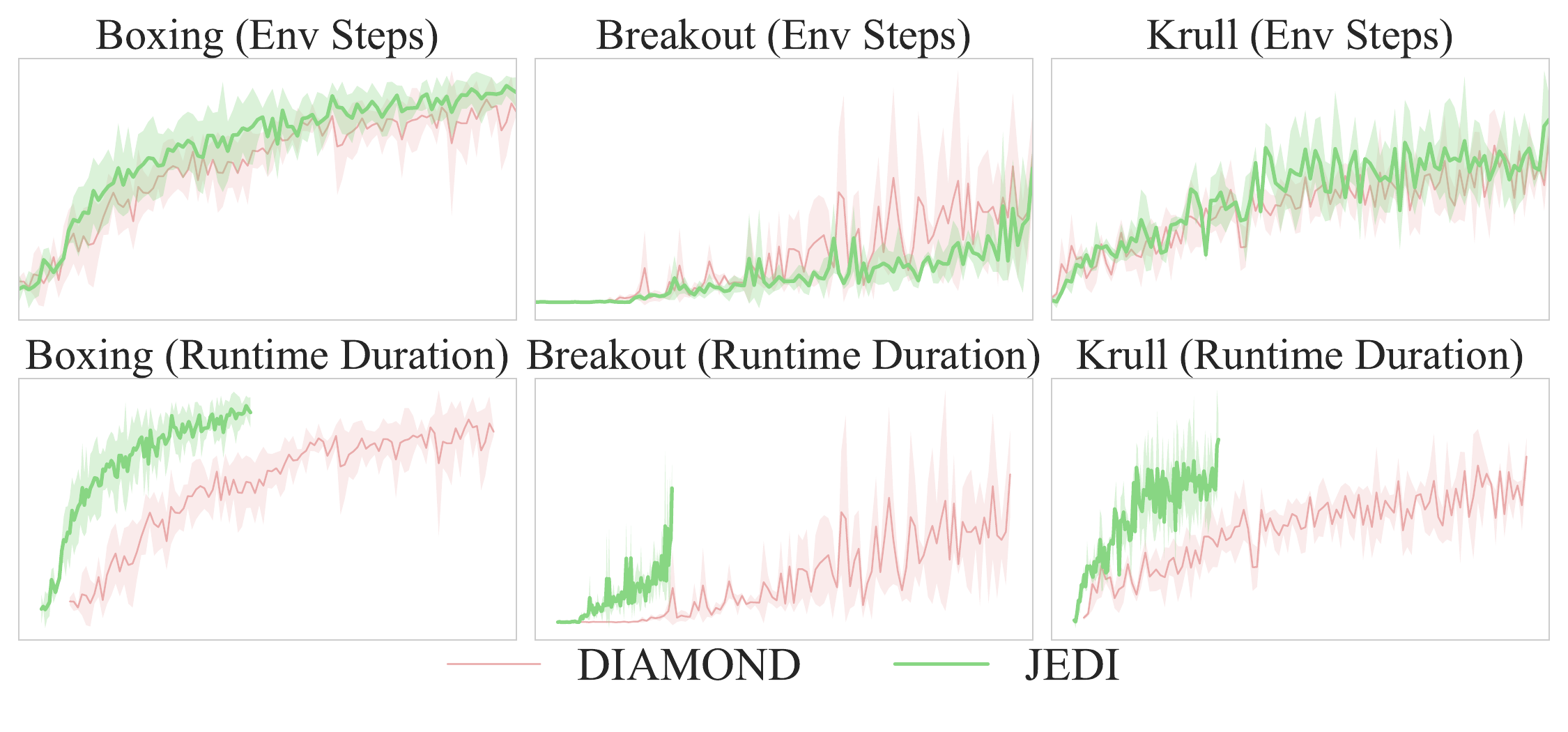}
  \caption{Training curves against environment steps (top) and actual runtime duration (bottom). JEDI's performance matches DIAMOND's while running 2$\times$ faster.}
  \label{fig:compute_and_runtime_comparison}
\end{figure}
\vspace*{-3mm}
\begin{figure}[H]
  \centering
  % First subfigure
    \centering
    \includegraphics[width=\textwidth]{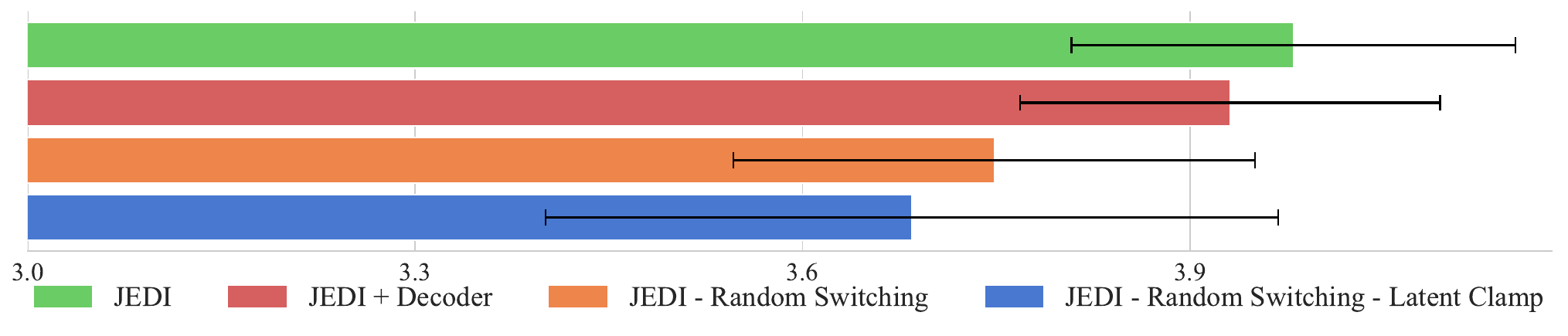}
  % \hfill
  % Second subfigure
  \caption{Technical Ablation study on 7 Agent-Optimal tasks . (1) JEDI w/o random switching of $\mathbf{E_\phi}$ and $\mathbf{D_\theta}$ output for following time step JEDI loss and w/o $\mathrm{tanh}$ scaled with a factor of 3 on the latent space, (2) JEDI w/o random switching, (3) JEDI w/ reconstruction decoder loss gradients, (4) JEDI. We tested the mechanisms we put in place for effective end-to-end latent diffusion and show that the clamping and random switching is the most effective, outperforming even gradients from decoder reconstruction.}
  \label{fig:old_ablation_study}
  \vspace{-10pt}
\end{figure}

\begin{figure}[h]
  \centering
  % First subfigure
  \begin{subfigure}[t]{0.4\textwidth}
    \centering
    \includegraphics[width=\textwidth]{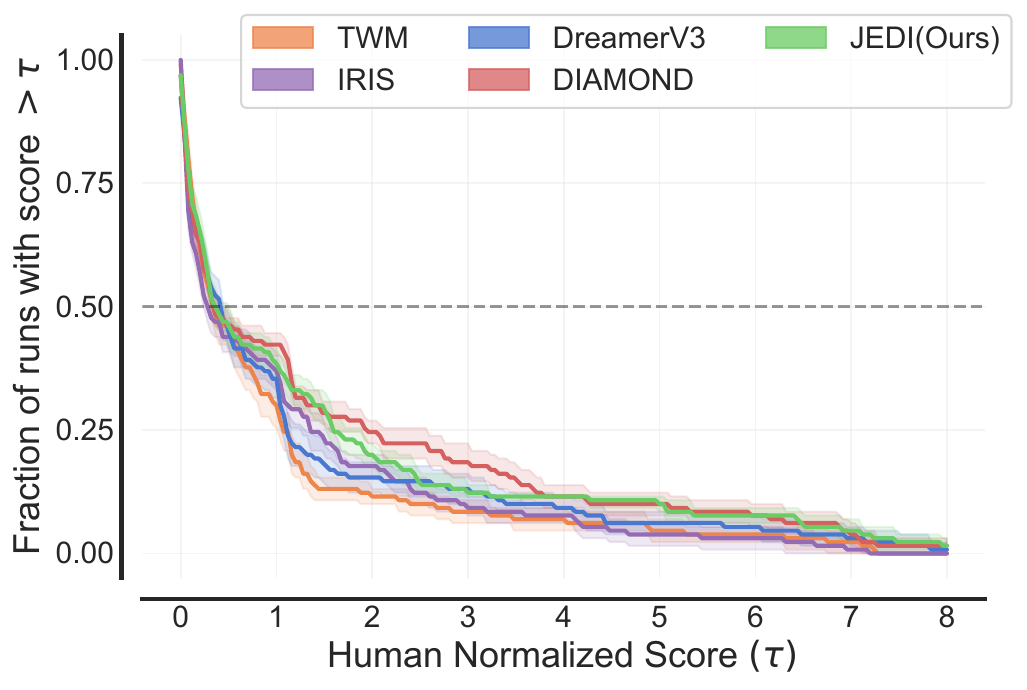}
    \caption{Original Performance Profile with $\tau\in[0,8]$ proposed by \citet{agarwal2021precipice}.}
    \label{fig:performance_profiles_orig}
  \end{subfigure}
  % Second subfigure
  \hspace{0.1\textwidth}
  \begin{subfigure}[t]{0.4\textwidth}
    \centering
    \includegraphics[width=\textwidth]{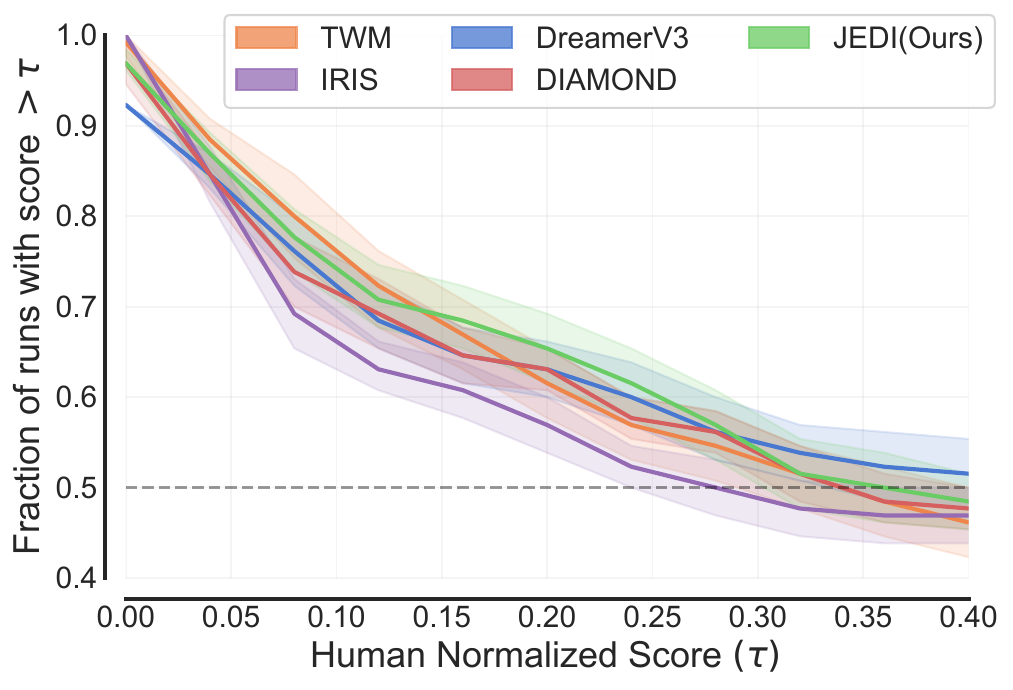}
    \caption{Truncated Performance Profile with $\tau\in[0,0.4]$ for better visualization.}
    \label{fig:performance_profiles_trunc}
  \end{subfigure}
  \caption{Performance profile plots. From (a), it appears that DIAMOND dominates from around $\tau\in[0.5,4]$, thereafter it matches JEDI in the range $\tau\in[4,8]$, suggesting that DIAMOND is SOTA. However, (a) fails to show that JEDI dominates over DIAMOND in the range $\tau\in[0,0.4]$. The large range of HNS conceals the significance of tasks with low HNS.}
  \label{fig:performance_profiles}
  
\end{figure}

\setlength{\tabcolsep}{2pt}
\begin{table}[h]                   
  \centering
  \footnotesize
  \caption{Atari100k overall performance. Blue font and red font refers to outperforming and underperforming to DIAMOND respectively. The tasks are sorted in decreasing HNS of the averaged agent level performance (\cref{sec:performance_bias}). Middle line indicates the split between the Agent-Optimal and Human-Optimal tasks respectively. Bold refers to the best result, underline refers to the runner-up. DIAMD = DIAMOND, DrmrV3 = DreamerV3. JEDI achieves SOTA optimality gap and runner-up performance in number of super-human games achieved and overall mean.}
  \label{tab:atari100k_FULL_NUMBERS}

    \begin{tabular}{
        l                                % Game
        rrrrrrrrrr                   % 13 numeric columns + HN
    }
\toprule
Game                 &  Random    &  Human              &  SimPLE    &  TWM                                  &  IRIS                        &  DreamerV3                   &  STORM                                &  DIAMOND            &  JEDI(Ours)                          \\
\midrule
Boxing               &  0.1       &  12.1               &  7.8       &  \textcolor{red}{77.5}                &  \textcolor{red}{70.1}       &  \textcolor{red}{73.8}       &  \textcolor{red}{79.7}                &  86.9               &  \textbf{\textcolor{blue}{91.6}}     \\
Krull                &  1598.0    &  2665.5             &  2204.8    &  \textcolor{red}{6349.2}              &  \textcolor{red}{6616.4}     &  \textcolor{red}{7921.5}     &  \textcolor{red}{8412.6}              &  \textbf{8610.1}    &  \textcolor{red}{8499.4}             \\
CrazyClimber         &  10780.5   &  35829.4            &  62583.6   &  \textcolor{red}{71820.4}             &  \textcolor{red}{59324.2}    &  \textcolor{red}{84880.0}    &  \textcolor{red}{66776.0}             &  \textbf{99167.8}   &  \textcolor{red}{64786.2}            \\
Gopher               &  257.6     &  2412.5             &  596.8     &  \textcolor{red}{1674.8}              &  \textcolor{red}{2236.1}     &  \textcolor{red}{5754.3}     &  \textbf{\textcolor{blue}{8239.6}}    &  5897.9             &  \textcolor{red}{4155.3}             \\
RoadRunner           &  11.5      &  7845.0             &  5640.6    &  \textcolor{red}{9109.0}              &  \textcolor{red}{9614.6}     &  \textcolor{red}{14995.0}    &  \textcolor{red}{17564.0}             &  \textbf{20673.2}   &  \textcolor{red}{14690.0}            \\
Jamesbond            &  29.0      &  302.8              &  100.5     &  \textcolor{red}{362.4}               &  \textcolor{blue}{462.7}     &  \textcolor{blue}{480.4}     &  \textbf{\textcolor{blue}{509.0}}     &  427.4              &  \textcolor{blue}{460.5}             \\
Assault              &  222.4     &  742.0              &  527.2     &  \textcolor{red}{682.6}               &  \textcolor{red}{1524.4}     &  \textcolor{red}{669.7}      &  \textcolor{red}{801.0}               &  \textbf{1526.4}    &  \textcolor{red}{1394.5}             \\
Breakout             &  1.7       &  30.5               &  16.4      &  \textcolor{red}{20.0}                &  \textcolor{red}{83.7}       &  \textcolor{red}{27.9}       &  \textcolor{red}{15.9}                &  132.5              &  \textbf{\textcolor{blue}{155.6}}    \\
KungFuMaster         &  258.5     &  22736.3            &  14862.5   &  \textcolor{blue}{24554.6}            &  \textcolor{blue}{21759.8}   &  \textcolor{blue}{24210.0}   &  \textbf{\textcolor{blue}{26182.0}}   &  18713.6            &  \textcolor{red}{12725.2}            \\
Pong                 &  -20.7     &  14.6               &  12.8      &  \textcolor{red}{18.8}                &  \textcolor{red}{14.6}       &  \textcolor{red}{18.9}       &  \textcolor{red}{11.3}                &  \textbf{20.4}      &  \textcolor{red}{18.2}               \\
Kangaroo             &  52.0      &  3035.0             &  51.2      &  \textcolor{red}{1240.0}              &  \textcolor{red}{838.2}      &  \textcolor{red}{3790.0}     &  \textcolor{red}{4208.0}              &  \textbf{5382.2}    &  \textcolor{red}{736.0}              \\
UpNDown              &  533.4     &  11693.2            &  3350.3    &  \textbf{\textcolor{blue}{15981.7}}   &  \textcolor{red}{3546.2}     &  \textcolor{blue}{7981.7}    &  \textcolor{blue}{7985.0}             &  3856.3             &  \textcolor{blue}{4191.1}            \\
Freeway              &  0.0       &  29.6               &  16.7      &  \textcolor{red}{24.3}                &  \textcolor{red}{31.1}       &  \textcolor{red}{0.0}        &  \textcolor{red}{0.0}                 &  \textbf{33.3}      &  \textcolor{red}{6.3}                \\
\midrule
BankHeist            &  14.2      &  753.1              &  34.2      &  \textcolor{blue}{466.7}              &  \textcolor{blue}{53.1}      &  \textcolor{blue}{622.7}     &  \textcolor{blue}{641.2}              &  19.7               &  \textbf{\textcolor{blue}{926.1}}    \\
DemonAttack          &  152.1     &  1971.0             &  208.1     &  \textcolor{blue}{350.2}              &  \textcolor{blue}{2034.4}    &  \textcolor{blue}{433.7}     &  \textcolor{red}{164.6}               &  288.1              &  \textbf{\textcolor{blue}{2047.2}}   \\
Hero                 &  1027.0    &  \textbf{30826.4}   &  2656.6    &  \textcolor{blue}{7254.0}             &  \textcolor{blue}{7037.4}    &  \textcolor{blue}{11145.2}   &  \textcolor{blue}{11044.3}            &  5621.8             &  \textcolor{blue}{7442.9}            \\
BattleZone           &  2360.0    &  \textbf{37187.5}   &  4031.2    &  \textcolor{blue}{5068.0}             &  \textcolor{blue}{13074.0}   &  \textcolor{blue}{12400.0}   &  \textcolor{blue}{13540.0}            &  4702.0             &  \textcolor{blue}{16084.0}           \\
Frostbite            &  65.2      &  \textbf{4334.7}    &  236.9     &  \textcolor{blue}{1475.6}             &  \textcolor{red}{259.1}      &  \textcolor{blue}{945.1}     &  \textcolor{blue}{1316.0}             &  274.1              &  \textcolor{red}{263.3}              \\
Qbert                &  163.9     &  \textbf{13455.0}   &  1288.8    &  \textcolor{red}{3330.8}              &  \textcolor{red}{745.7}      &  \textcolor{red}{3291.3}     &  \textcolor{blue}{4522.5}             &  4499.3             &  \textcolor{red}{3606.1}             \\
MsPacman             &  307.3     &  \textbf{6951.6}    &  1480.0    &  \textcolor{red}{1588.4}              &  \textcolor{red}{999.1}      &  \textcolor{red}{1358.4}     &  \textcolor{blue}{2673.5}             &  1958.2             &  \textcolor{red}{1075.9}             \\
Asterix              &  210.0     &  \textbf{8503.3}    &  1128.3    &  \textcolor{red}{1116.6}              &  \textcolor{red}{853.6}      &  \textcolor{red}{946.4}      &  \textcolor{red}{1028.0}              &  3698.5             &  \textcolor{red}{2331.0}             \\
ChopperCommand       &  811.0     &  \textbf{7387.8}    &  979.4     &  \textcolor{blue}{1697.4}             &  \textcolor{blue}{1565.0}    &  \textcolor{red}{411.7}      &  \textcolor{blue}{1888.0}             &  1369.8             &  \textcolor{blue}{1392.4}            \\
Amidar               &  5.8       &  \textbf{1719.5}    &  74.3      &  \textcolor{red}{121.8}               &  \textcolor{red}{143.0}      &  \textcolor{red}{141.2}      &  \textcolor{red}{204.8}               &  225.8              &  \textcolor{red}{114.3}              \\
Alien                &  227.8     &  \textbf{7127.7}    &  616.9     &  \textcolor{red}{674.6}               &  \textcolor{red}{420.0}      &  \textcolor{blue}{1078.9}    &  \textcolor{blue}{983.6}              &  744.1              &  \textcolor{blue}{1090.1}            \\
PrivateEye           &  24.9      &  \textbf{69571.3}   &  35.0      &  \textcolor{red}{86.6}                &  \textcolor{red}{100.0}      &  \textcolor{blue}{1081.4}    &  \textcolor{blue}{7781.0}             &  114.3              &  \textcolor{red}{95.2}               \\
Seaquest             &  68.4      &  \textbf{42054.7}   &  683.3     &  \textcolor{blue}{774.4}              &  \textcolor{blue}{661.3}     &  \textcolor{blue}{610.0}     &  \textcolor{red}{525.2}               &  551.2              &  \textcolor{blue}{1103.7}            \\
\midrule
\#Superhuman (↑)     &  0         &  N/A                &  1         &  8                                    &  \underline{10}              &  9                           &  9                                    &  \textbf{11}        &  \textbf{11}                                  \\
Mean (↑)             &  0.000     &  1.000              &  0.332     &  0.956                                &  1.046                       &  1.134                       &  1.222                                &  \textbf{1.459}     &  \underline{1.361}                               \\
Median (↑)           &  0.000     &  1.000              &  0.134     &  \textbf{0.505}                       &  0.289                       &  \underline{0.503}           &  0.425                                &  0.373              &  0.361                               \\
IQM (↑)              &  N/A       &  N/A                &  0.130     &  0.459                                &  0.501                       &  0.504                       &  0.561                                &  \textbf{0.641}     &  \underline{0.609}                               \\
Optimality Gap (↓)   &  N/A       &  N/A                &  0.729     &  0.513                                &  0.512                       &  0.500                       &  \textbf{0.472}                       &  \underline{0.480}  &  \underline{0.480}                               \\
      \bottomrule
    \end{tabular}
\end{table}

\begin{table}[ht]
  \centering
  % \vspace{-8pt}
  % \scriptsize
  \caption{Aggregate Atari100k performance of all methods. Bold = best, underline = runner-up. Mean, Median, and Interquantile Mean (IQM) is with reference to Human Normalized Scores (↑); Optimality Gap (↓) is the overall gap to human-level performance \citep{agarwal2021precipice}. DIAMD = DIAMOND, DrmrV3 = DreamerV3. JEDI achieves SOTA results on number of Super-Human tasks, mean and Optimality Gap, and runner-up on IQM.}
  \label{tab:atari100k_aggregate_statistics}

  % \resizebox{\columnwidth}{!}{
      \begin{tabular}{l *{6}{r}}
        \toprule
        \textbf{Human-Optimal} & IRIS & TWM & DrmrV3 & STORM & DIAMD & JEDI \\
        \midrule
            \#Superhuman (↑)     &  \underline{1}        &  0                 &  0                 &  0                 &  0                 &  \textbf{2}        \\
            Mean (↑)             &  0.162    &  0.166             &  0.190             &  0.238             &  0.127             &  \textbf{0.297}    \\
            IQM (↑)              &  0.072    &  0.112             &  0.122             &  \textbf{0.184}    &  0.091             &  \underline{0.156}             \\
            Optimality Gap (↓)   &  0.852    &  0.840             &  0.823             &  0.766             &  0.873             &  \textbf{0.740}    \\
        \bottomrule
      \end{tabular}
        % }
            % Median (↑)           &  0.078    &  0.109             &  \underline{0.155}             &  \textbf{0.164}    &  0.075             &  0.125             \\
        
  % \resizebox{\columnwidth}{!}{
      \begin{tabular}{l *{6}{r}}
        \toprule
        \textbf{Agent-Optimal} & IRIS & TWM & DrmrV3 & STORM & DIAMD & JEDI \\
        \midrule
            \#Superhuman (↑)     &  \underline{9}        &  8                 &  \underline{9}           &  \underline{9}                 &  \textbf{11}        &  \underline{9}                 \\
            Mean (↑)             &  1.930    &  1.746             &  2.078       &  2.206             &  \textbf{2.791}     &  \underline{2.425}             \\
            IQM (↑)              &  1.461    &  1.051             &  1.303       &  1.486             &  \textbf{2.150}     &  \underline{1.640}             \\
            Optimality Gap (↓)   &  \underline{0.172}    &  0.185             &  0.176       &  0.177             &  \textbf{0.087}     &  0.219             \\
        \bottomrule
      \end{tabular}
            % }
            % Median (↑)           &  1.226    &  1.119             &  1.253       &  1.393             &  \textbf{2.510}     &  \underline{1.809}             \\

\end{table}

\newpage
% \subsection{Raw Scores for Average Agent Performance}
% \label{sec:raw_scores_for_agent_level_performance}
% The agent-level performance we used as reference to derive the Agent-Optimal tasks and Human-Optimal tasks was derived from the averaged raw scores of the 4 baselines \citep{robine2023TWM, micheli2022iris, hafner2023Dreamerv3, zhang2023storm} selected in \cref{sec:performance_bias}. Below is the agent-level performance raw score for each game.
\begin{table}[h]
    \centering
        \begin{tabular}{l|l|l|l}
        \toprule
            Agent-Optimal Tasks      & Raw Scores      & Human-Optimal Tasks                & Raw Scores \\
        \midrule
        \textit{Boxing}              & 76.325          & \textit{BankHeist}           & 452.5        \\
        \textit{Krull}               & 7290.05         & \textit{DemonAttack}         & 713.05        \\
        \textit{CrazyClimber}        & 73777.65         & \textit{Hero}                & 9124.175        \\
        \textit{Gopher}              & 3970.125         & \textit{BattleZone}          & 10983       \\
        \textit{RoadRunner   }       & 12963.15         & \textit{Frostbite}           & 989.925        \\
        \textit{Jamesbond    }       & 444.775         & \textit{Qbert}               & 3001        \\
        \textit{Assault}             & 928.5            & \textit{MsPacman}            & 1647        \\
        \textit{Breakout}            & 37.65            & \textit{Asterix}             & 982.55       \\
        \textit{KungFuMaster}        & 23479.1         & \textit{ChopperCommand}      & 1392.6     \\
        \textit{Pong}                & 15.675         & \textit{Amidar}              & 152.15       \\
        \textit{Kangaroo}            & 2596.05         & \textit{Alien}               & 759.3       \\
        \textit{UpNDown }            & 9186.725         & \textit{PrivateEye}          & 2212.4       \\
        \textit{Freeway}             & 22.225        & \textit{Seaquest}            & 644.725       \\
        \bottomrule
        \end{tabular}
    \caption{Raw scores for Average Agent Level Performance. The Average Agent is defined in \cref{sec:performance_bias}}
    \label{tab:agent_level_performance_raw_scores}
\end{table}

\newpage
%%%%%%%%%%%%%%%%%%%%%%%%%%%%%%%%%%%%%%%%%%%%%%%%%%%%%%%%%%%%%%%%%%%%%%%%%%%%%%%
%%%%%%%%%%%%%%%%%%%%%%%%%%%%%%%%%%%%%%%%%%%%%%%%%%%%%%%%%%%%%%%%%%%%%%%%%%%%%%%

\end{document}